\documentclass{article}

\usepackage[preprint]{neurips_2025}

\usepackage{amsthm}

\usepackage{microtype}
\usepackage{graphicx}
\usepackage{subcaption}
\usepackage{booktabs} 
\usepackage{tikz} 

\usepackage{hyperref}
\usepackage{algorithmic}
\usepackage{amsmath,amssymb,amsthm} 
\usepackage{amsfonts,amsmath}
\usepackage[linesnumbered,ruled]{algorithm2e}
\usepackage{psfrag,xspace}
\usepackage{color,etoolbox}
\usepackage{wasysym}
\usepackage[export]{adjustbox}
\usepackage[T1]{fontenc} 

\allowdisplaybreaks

\usepackage[english]{babel}

\DeclareSymbolFont{bbold}{U}{bbold}{m}{n}
\DeclareSymbolFontAlphabet{\mathbbold}{bbold}

\usepackage{cancel}

\theoremstyle{definition}

\theoremstyle{remark}

\usepackage{comment}

\usepackage[utf8]{inputenc} 
\usepackage[T1]{fontenc}    
\usepackage{hyperref}       
\usepackage{url}            
\usepackage{booktabs}       
\usepackage{amsfonts}       
\usepackage{nicefrac}       
\usepackage{microtype}      
\usepackage{xcolor}         

\title{Failure Modes of LLMs for Causal Reasoning on Narratives}

\author{Khurram Yamin\footnote{Equal contribution}}

\author{%
  Khurram Yamin\\
  \texttt{kyamin@andrew.cmu.edu} \\
  \texttt{Carnegie Mellon}\\
  \And
  Shantanu Gupta \\
  \texttt{shantang@andrew.cmu.edu} \\
  \texttt{Carnegie Mellon} \\
  \And
  Gaurav Ghosal \\
  \texttt{gghosal@andrew.cmu.edu} \\
  \texttt{Carnegie Mellon}
  \And
  Zachary Lipton \\
  \texttt{zlipton@andrew.cmu.edu} \\
  \texttt{Carnegie Mellon}
  \And
  Bryan Wilder \\
  \texttt{bwilder@andrew.cmu.edu} \\
  \texttt{Carnegie Mellon} \\
}

\begin{document}

\maketitle

\begin{abstract}
The ability to robustly identify causal relationships is essential for autonomous decision-making and adaptation to novel scenarios. However, accurately inferring causal structure requires integrating both world knowledge and abstract logical reasoning. In this work, we investigate the interaction between these two capabilities through the representative task of causal reasoning over narratives. Through controlled synthetic, semi-synthetic and real-world experiments, we find that state-of-the-art large language models (LLMs) often rely on superficial heuristics—for example, inferring causality from event order or recalling memorized world knowledge without attending to context. Furthermore, we show that simple reformulations of the task can elicit more robust reasoning behavior. Our evaluation spans a range of causal structures, from linear chains to complex graphs involving colliders and forks. These findings uncover systematic patterns in how LLMs perform causal reasoning and lay the groundwork for developing methods that better align LLM behavior with principled causal inference.

\end{abstract}

\section{Introduction}

Large Language Models (LLMs) have achieved promising performance across a range of tasks, as a result of their ability to absorb rich world knowledge from large-scale unsupervised data. As such, there is rising interest in deploying them in autonomous and agentic settings. However, a key prerequisite for autonomous decision making is causal reasoning: agents must be able to move beyond relying on associations and be capable of robustly determining the consequences of their actions. Does this capability arise naturally from large-scale pretraining?

Robust causal reasoning is particularly challenging as it relies on a combination of knowledge and reasoning capabilities. Unlike mathematical reasoning benchmarks \citep{cobbe2021trainingverifierssolvemath} -- which draw from a relatively constrained set of problem solving strategies and results -- arriving at correct causal inferences often requires leveraging domain-specific knowledge about the events or variables involved. On the other hand, models must go beyond blindly retrieving memorized associations and knowledge to identify the correct relationships under atypical or counter-intuitive settings. World knowledge and reasoning must be carefully balanced for robust causal reasoning: incorporating some memorized world knowledge can be helpful, but over-reliance can lead to errors in atypical settings.


Prior works have primarily studied reasoning and world-knowledge of LLMs separately. For example, benchmarks on mathematical reasoning or coding typically study these capabilities in isolation -- with minimal external world knowledge needed to solve problems. On the other hand, benchmarks for knowledge intensive tasks can generally be solved by simple retrieval of memorized knowledge. Thus, the interplay of knowledge retrieval and reasoning (and potential conflicts between them) remains understudied.

In this work, we study this interplay in the setting of causal reasoning over textual narratives. We pay particular attention to (a) \emph{long sequences of events} (as are likely to arise in autonomous LLM settings) and (b) \emph{atypical settings} where typical ``common-sense" causal relationships may not hold. Concretely, 
we start by considering settings where there is an (unknown) 
underlying causal chain graph of the form $V_1 \rightarrow V_2 \rightarrow \hdots \rightarrow V_N$, where each node $V_i$ has some semantic meaning (e.g., smoking), 
that is verbalized in the form of a (realistic) narrative $N$.
For a given narrative $N$,
we consider two causal reasoning tasks: 
(1) Does $V_i$ have a causal effect on $V_j$? and 
(2) Given the node identities $(V_1, \hdots V_N)$, 
construct a causal chain graph that is faithful to the narrative.
While these tasks do not encompass all aspects of causal reasoning 
(e.g., counterfactual reasoning),
they are important primitives for successful causal reasoning. We then extend our analysis to more complex underlying graph structures including colliders and forks.

Our primary contribution is to characterize when large language models are able to robustly reason under these causal structures. On the one hand, we find that LLMs display  distinctive failure modes related to interference between reasoning and world knowledge in causal inference.  Firstly, we show that LLMs are influenced heavily by a prior that causes are likely to appear before effects in a narrative. We observe that when the narrative is constructed in the reverse topological order of the causal chain (i.e., the edge $V_i \rightarrow V_{i+1}$ is narrated \emph{before} $V_{i-1} \rightarrow V_{i}$), the performance of the LLM suffers as it often assigns the cause to an earlier event and the effect to a later event in the narrative. Secondly, we show that LLMs use their
parametric causal knowledge
(i.e., if an event typically causes another event)
as a shortcut
to answer causal questions.
Thus, when the cause-and-effect pairs
implied by the narrative conflict
with the parametric knowledge,
the LLM often ignores the specifics of the narrative and
defaults to its parametric knowledge. Neither prompting with Chain of Thought (CoT) \citep{wei2022chain} nor In-Context Learning alleviates these failures.

However, LLMs are much less impacted by variation in the reasoning difficulty of the task when the prompting scheme explicitly isolates reasoning and world knowledge. First, we find that asking the LLM to extract the entire causal graph implied by the narrative results in a high degree of success at correctly ordering individual events, largely avoiding both failure modes described above. However, these benefits dissipate if the model is prompted to use the extracted graph alongside the narrative. Second, LLMs exhibit only slight performance degradation when reasoning over narratives that display more complex graph structures than chains, for example forks or colliders. Third, while LLMs often struggle with longer narratives containing more events, this failure is also substantially mitigated by asking the LLM to just extract a graph. All together, our results paint a more nuanced picture of LLMs' causal capabilities than simple success or failure and suggest that future development should focus on isolating and then composing LLMs' strengths at reasoning and world knowledge in order to avoid conflicts between them. We validate our findings across both controlled synthetic (LLM generated) narratives and real-world articles from \textit{CauseNet} \cite{heindorf2020causenet}.

Ultimately, our work presents a comprehensive study of causal reasoning in LLMs. Our findings reveal that current state-of-the-art models can fail to reliably perform simple causal reasoning tasks on narratives. We distill two shortcuts that underlie this unreliability: over reliance on positional biases and parametric knowledge. On the other hand, we find that incorporating an \textit{explicit} causal graph identification task can elicit significantly better reasoning behavior. 

\section{Related Works}
\paragraph{Causal Reasoning in Large Language Models}  \citet{jin2023cladder} develop a benchmark for testing causal reasoning in LLMs given causal graphs, finding that language models can struggle with the task. However, the queries examined in \citet{jin2023cladder} require probability calculations, potentially conflating causal reasoning and arithmetic failures.   \citet{tan-etal-2022-causal} shows the capability of a neural network trained on news data to label causal structures in individual sentences.  \citet{joshi2024llmspronefallaciescausal} chronicles failure modes in textual, but non-narrative form data (e.g. text formulaically written as Event 1 Causes Event 2 Causes Event 3 Causes Event 4). Our paper expands upon such a line of work by testing the LLM's abilities in both real and synthetic texts that would plausibly be seen in everyday life.  Another contrasting work, \citet{jin2024can}, uses only statistical language indicating event correlations as input.

\citep{gordon-etal-2012-semeval,joshi2024coldcausalreasoningclosed,ho2023wikiwhy,zhang2023understanding,wang-etal-2023-cola, ashwani2024causeeffectlargelanguage} study causal reasoning ability as it relates to inferring causal relations based on "common sense". In such common-sense based settings, it is straightforward for models to simply rely on memorized knowledge from pretraining and achieve good performance, without leveraging any more general causal reasoning capabilities. Our work seeks to disentangle this general causal reasoning ability by specifically testing cases where causal relationships may contradict common-sense knowledge. This serves as a more robust measurement of the causal reasoning capabilities in unfamiliar and atypical scenarios. Empirically, we show that models struggle significantly in adapting to unfamiliar causal relations. 

Another important distinction of our work is the focus on longer-form narratives. Existing works such as \citep{gordon-etal-2012-semeval,zečević2023causalparrotslargelanguage, ho2023wikiwhy, frohberg2022crassnoveldataset, li2023counterfactualreasoningtestinglanguage, gao2023chatgptgoodcausalreasoner} primarily examine short-form questions about a single causal relationship. On the other hand, our work examines longer and more complex sequences of events. Moreover, in contrast to domain-specific question banks such as Intuitive Physics studied in \citep{zečević2023causalparrotslargelanguage}, our narratives examine a more diverse range of topics (as illustrated by the sample narratives presented). As a result, our dataset provides a more realistic and diverse examination of LLM causal reasoning capabilities than prior works. As such, our work is unique in that we are the first paper to analyze non-common sense based causal reasoning in narratives that use everyday language.

\paragraph{Causal Story Generation}
\citet{kıcıman2024causalreasoninglargelanguage} shows that LLMs have strong abilities to generate causal texts. \citet{ammanabrolu2020automatedstorytellingcausalcommonsense} introduces soft causal relations—causal constraints that match what readers expect—and uses commonsense inferences to bridge high-level plot points, resulting in more coherent narratives that align with everyday causal expectations. \citet{tian2021hypogenhyperbolegenerationcommonsense} 
 contributes by employing counterfactual knowledge to generate hyperboles, making story generation more realistic. \citet{li2022rmcsmallis2simplified} shows that asking a model to explain a cause or effect by generating new text conflates language generation with prediction; instead, their approach asks the model to simply indicate the sentence number representing the cause or effect, leading to stories that better respect causal relations.  For our synthetic text generation, we focus on creating narratives that are extremely explicit and simple. In contrast to \citet{ammanabrolu2020automatedstorytellingcausalcommonsense} and \citet{li2022rmcsmallis2simplified}'s approach of bridging events using commonsense, our narrative scheme already embeds explicit causal language between events that are causally related so that no inference or common-sense reasoning is required from the reader to reason about causality. Furthermore, our experiments often intentionally contradict common-sense parametric knowledge to check the model's ability to solely rely on the self-contained narrative. Similarly, regarding \citet{tian2021hypogenhyperbolegenerationcommonsense}, we opted to avoid abstract language structures that might confuse even human readers.

\section{Experiments with Synthetic Data}\label{sec:synthetic-data}

\subsection{Setting}
\paragraph{Synthetic Narrative Generation} In our synthetic experiments, we use three leading LLMs:
OpenAI's GPT-4o \citep{openai2024gpt4technicalreport}, Anthropic's Claude 
3.5 Sonnet \citep{Claude_35}, and the open source LLama 3.1 8b \citep{grattafiori2024llama3herdmodels}. While we focus on GPT-4o in the main text, results from other models are in the Appendix.
The purpose of our synthetic setup is to carefully control the conditions under which the LLMs are tested. 
In terms of the general setup of our fully synthetic experiments, 
we first use the LLM to generate events 
(which are real world phenomena like \emph{rain} or \emph{plants growing}). 
Then these events are linked together into a chain graph $G$ 
that acts as the causal ground truth (eg \emph{rain} $\rightarrow$ \emph{plants growing}). 
The LLM is given $G$ and asked to create a narrative that stays faithful to the causal relationships in $G$. These narratives are checked by researchers to ensure consistency with their base causal graphs. More specifically, when constructing the dataset, we asked researchers (select authors who were blind to the true underlying graph) to reconstruct the causal chains given just the narratives, and 98  percent of the time (out of 100 random samples), the humans were able to find the unique correct causal ordering. Roughly 2500 narrative samples were generated. To ensure a variety of events go into the narratives, we generate 100 to 1000 distinct events at a time and randomly pick the small number needed for narrative construction (all narratives are in supplementary files and select narratives are in the Appendix). 

Providing only the narrative as input (and not $G$), 
we then ask the LLM to find $G'$,  
the predicted underlying causal structure 
expressed by the narrative. In other words, the LLM is asked to output a causal graph that it thinks embodies the relationships in the narrative.  
Next, a series of causal questions is created
by randomly sampling $10$ tuples of events from $G$
and asking the LLM whether an event in 
the tuple causes the other based on the narrative and/or $G'$.     

\paragraph{Prompting Strategies} We evaluate five prompting styles for causal reasoning where the names in italics represent those used in the legends of figures: \textbf{Standard QA Prompting} (\textit{Standard}), where the model is simply asked to identify the causal relation between two narrative events; \textbf{Chain-of-Thought} (\textit{CoT}), which instructs the model to articulate step-by-step reasoning before answering; \textbf{In-Context Learning} (\textit{In-Context}), which precedes the query with illustrative question–answer examples; \textbf{Explicit Causal Graph Extraction} (\textit{Graph}), which asks the model to generate an entire causal graph $G'$ over all events and assesses whether the ordering of the target pair is correct; \textbf{Narrative-Augmented Graph Extraction} (\textit{Narr-Graph}), which first elicits $G'$ and then supplies both $G'$ and the original narrative for joint reasoning about the causal pair. Exact prompts are in Appendix~\ref{sec:apdx-synthetic}.

\begin{figure*}
    
    \begin{subfigure}[b]{0.45\textwidth}
\centering
\includegraphics[scale=0.32]{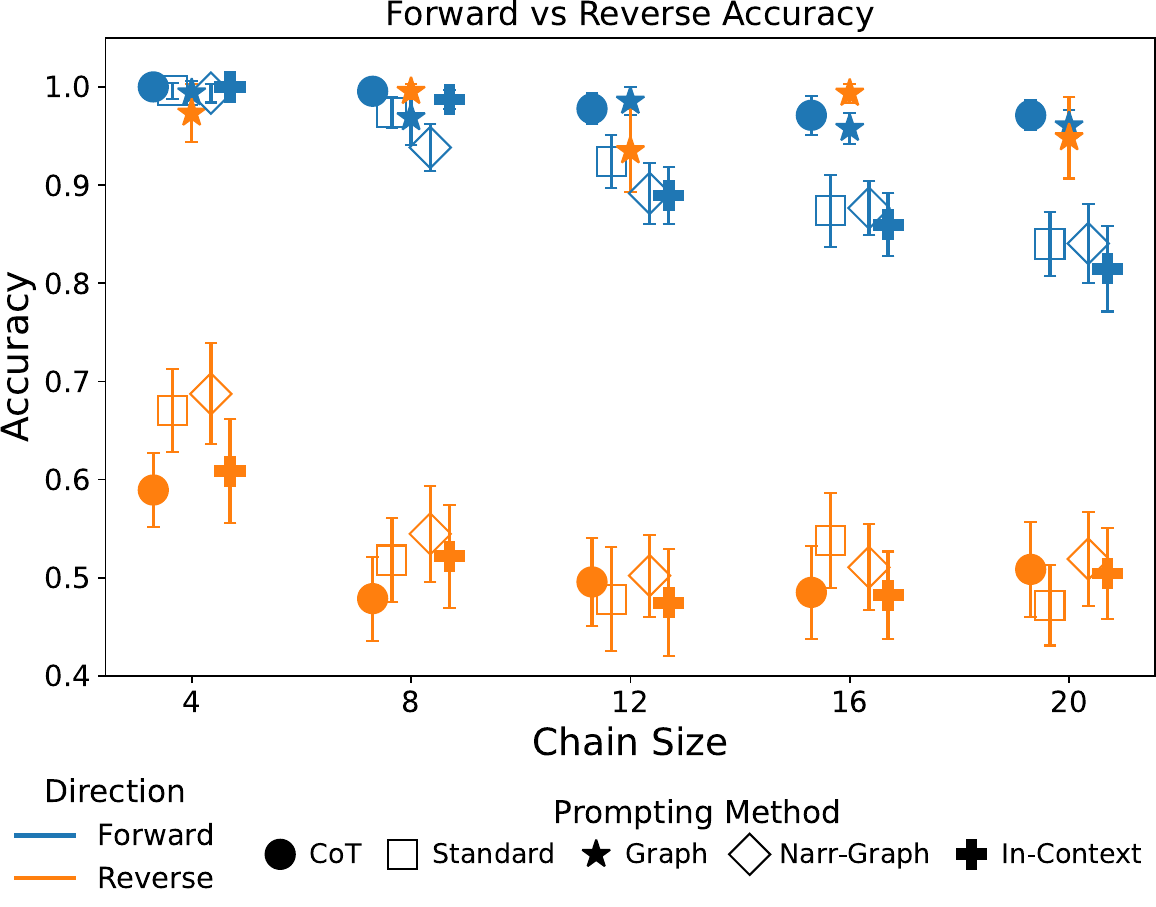}
\label{fig:place}
\end{subfigure}
\hspace{10px}
\begin{subfigure}[b]{0.45\textwidth}
\centering
\includegraphics[scale=0.32]{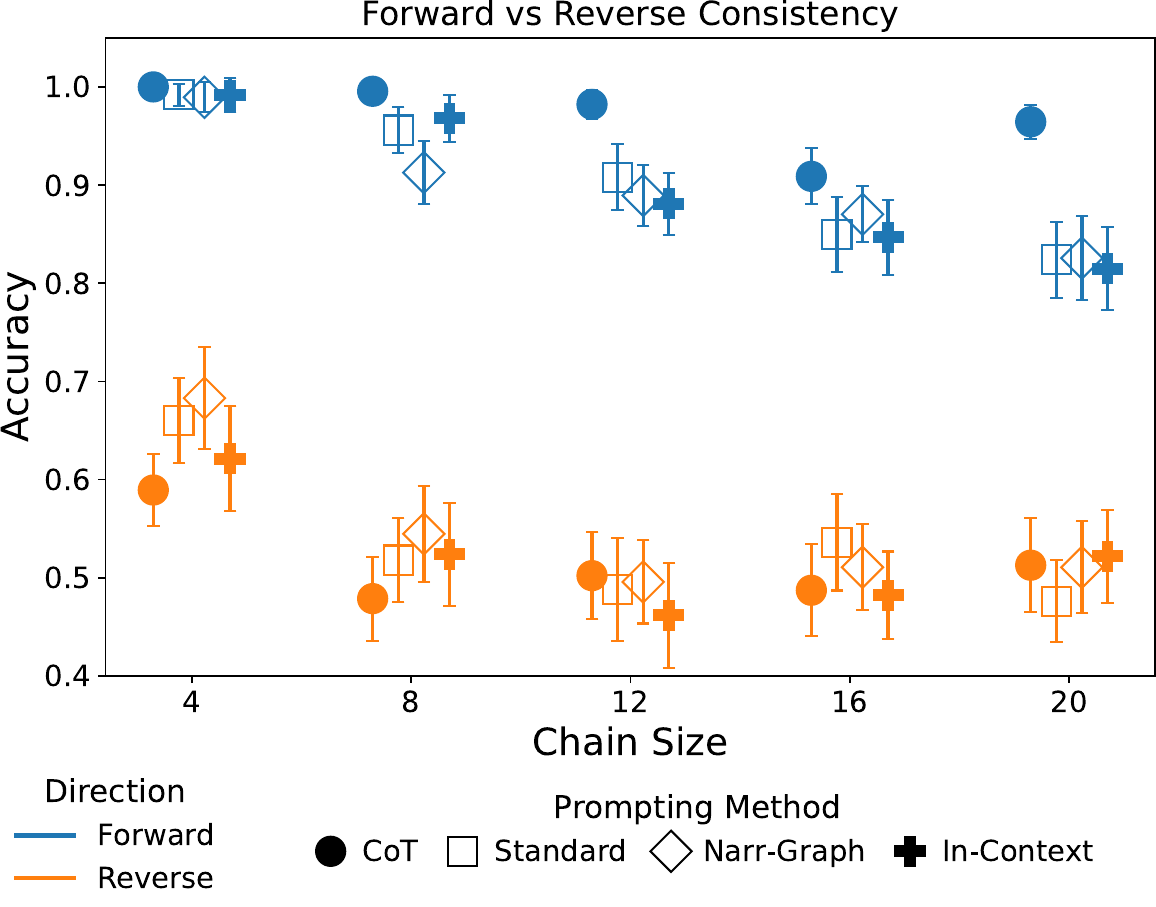}
\label{fig:place1}
\end{subfigure}

    \caption{GPT-4o Test of the LLM's ability to reason on narratives written in the Forward and Reverse topological orientations. Chain size is the number of nodes in ground truth $G$. The "Graph" prompting method uses only the extracted graph $G'$ to reason,  "Narr-Graph" uses both the narrative and extracted graph, and "Standard, CoT, In-Context" all use only the narrative.
    Accuracy measures LLM answer agreement with $G$, and consistency measures agreement with $G'$. The points in the graph are represented with a slight horizontal stagger around the relevant chain sizes (4,8,12 etc)  for ease of visual understanding. 
    We show a 95$\%$ CI.}
    \label{fig:forward}
\end{figure*}

\subsection{Impact of Event Ordering}
Our experiments show that LLMs rely on the 
ordering in which the events are verbalized
in a narrative when determining causal relationships. 
To investigate this, 
we started with randomly generated events 
that were used to make a ground truth graph $G$. 
During the creation of the narrative, 
we specified that the LLM either places the events 
in 
(1) the order that matches the topological causal ordering of the graph
(e.g., if event $A$ (indirectly or directly) causes $B$, 
then event $A$ is mentioned before $B$ in the narrative), 
or 
(2) a way that runs opposite to the causal ordering 
(event $B$ would be mentioned before event $A$ in the narrative
even though $A$ (directly or indirectly) causes $B$). 
We refer to these as the \emph{Forward} 
and \emph{Reverse} topological ordering, respectively. 
As an example, the following is a GPT-4o 
generated \emph{Reverse} topological narrative 
for the causal chain: \emph{Art exhibition}$\rightarrow$ \emph{Wine tasting} $\rightarrow$ \emph{Charity fundraiser}:
\begin{quote}
    The \emph{charity fundraiser} was made possible because of the successful \emph{wine tasting event} that attracted numerous generous patrons. The \emph{wine tasting} was organized as a result of the \emph{art exhibition} drawing in a sophisticated audience interested in cultural experiences.
\end{quote}
Each edge in the narrative is verbalized 
in the opposite order to its place in the causal chain. 
All narratives can be found in the linked code. 

\paragraph{LLMs Rely on Event Ordering Across Prompting Strategies} As shown in Figure~\ref{fig:forward} (left), 
in the \emph{Forward} direction,
standard QA, CoT, and In-Context prompts  all perform very well. This is in contrast to the \emph{Reverse} orientation when we look at the performance of the standard QA, COT, and In-Context prompts. From this plot, we can see
that naive COT and In-Context prompting do not seem to significantly boost accuracy under our conditions. 
Perhaps more interestingly, we find that the way the LLM answers questions using the narratives is not always consistent with the causal graph $G'$ that the LLM builds when asked to predict the underlying graph structure (see consistency plot in right side of Figure \ref{fig:forward},
where consistency measures agreement between the answers of the LLM and $G'$). In the \emph{Reverse} orientation, answers given by the extracted causal graph G' and the previously discussed prompting strategies seem to differ greatly. Additionally, the trend of those prompting strategies on the consistency plot for the \emph{Forward} orientation narratives (comparing performance to $G'$)  
mirrors their trend on the accuracy plot which compares performance to ground truth $G$ (left side). 

\paragraph{Explicit Causal Graph Extraction Avoids Shortcuts}This led us to test the accuracy of only using the extracted graph $G'$ to answer causal questions. In this case, once $G'$ is extracted by the LLM, it is
not given to the LLM again
to answer questions (but rather used directly).
We found that this strategy did significantly 
better in the \emph{Reverse} direction
than the other prompting strategies. Surprisingly, using $G'$ in the \emph{Reverse} direction narratives to answer causal questions did as well as using $G'$ in the \emph{Forward} direction narratives. Next, we tried prompting using the narrative and $G'$ (the LLM is given $G'$ in this case in the prompt). This technique could be thought of as a type of CoT prompting strategy. However, in the \emph{Reverse} direction narratives, the increase in accuracy achieved by only using $G'$ completely dissipates. We conjecture that the process of building the extracted Causal Graph $G'$ forces the LLM to engage in long term reasoning instead of using the simple shortcut, but when the narrative is again provided - the LLM defaults back to the shortcut.

\begin{figure*}

    \begin{subfigure}[b]{0.45\textwidth}
\centering
\includegraphics[scale=0.32]{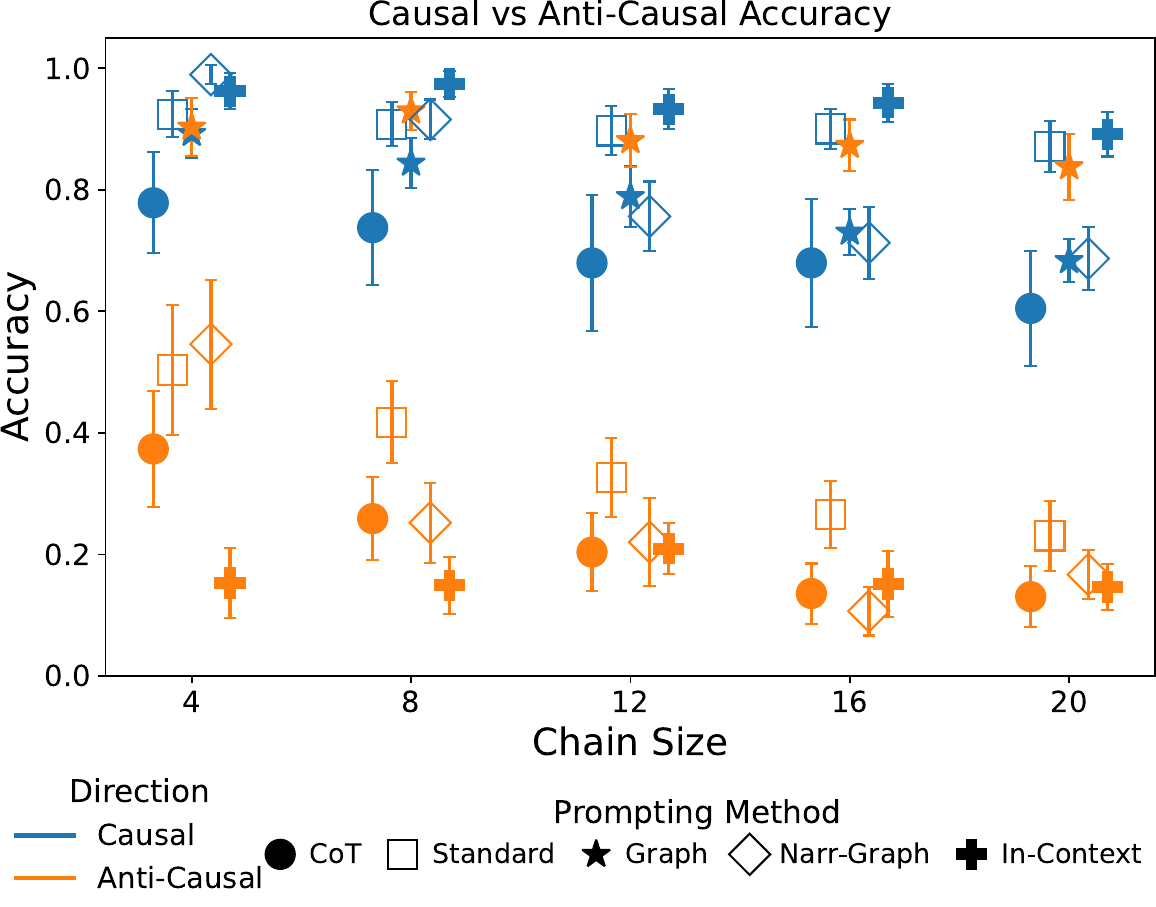}
\label{fig:place}
\end{subfigure}
\hspace{10px}
\begin{subfigure}[b]{0.45\textwidth}
\centering
\includegraphics[scale=0.32]{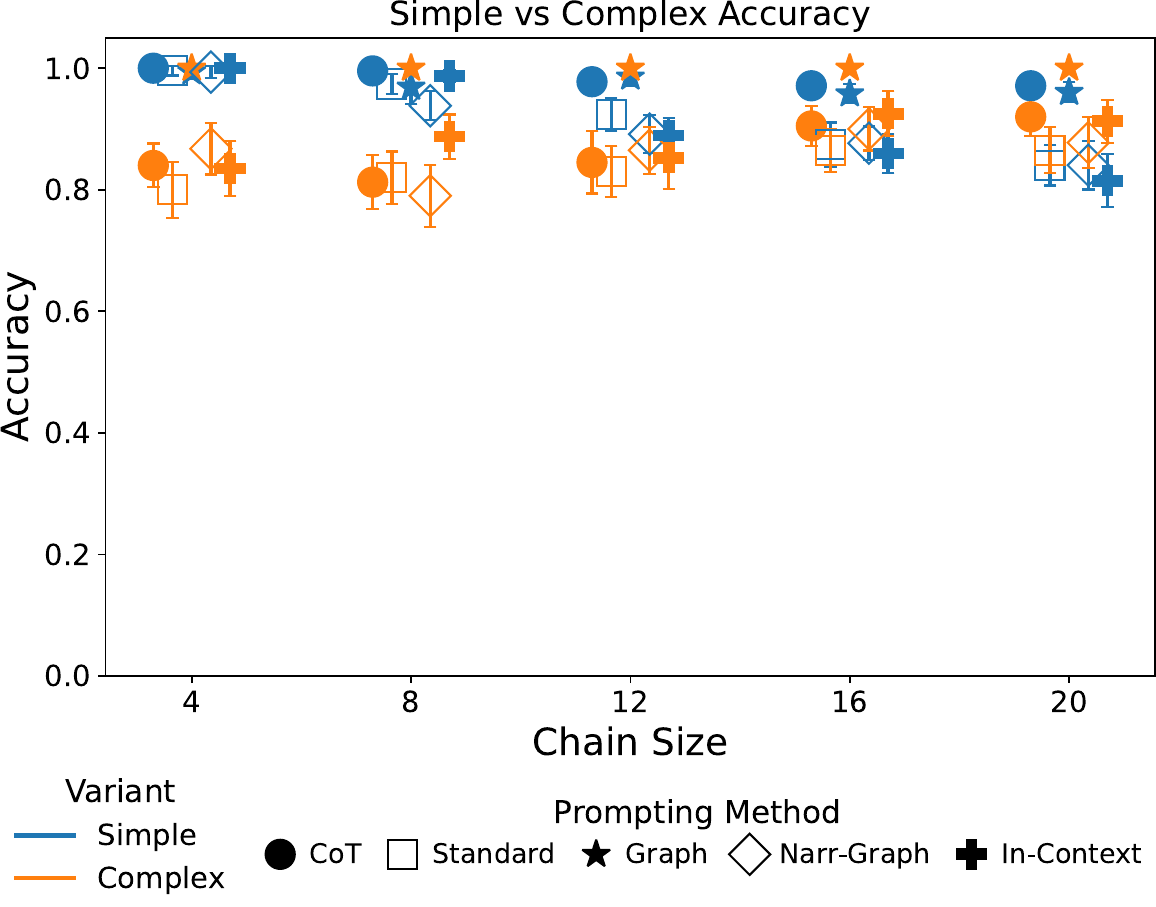}
\label{fig:place1}
\end{subfigure}

    \caption{ (Left) GPT-4o test of the LLM's ability to reason on narratives that agree with parametric knowledge (Causal) and disagree with parametric knowledge (Anti-Causal). (Right) GPT-4o test of the LLM's ability to reason on narratives generated from Complex graphs as opposed to Simple chain graphs. Label descriptions for both images match those of Figure \ref{fig:forward} and 95 $\%$ CI is shown. 
    \label{fig:param}}

\end{figure*}

\subsection{Impact of Parametric Knowledge (In)consistency}

\paragraph{Experimental Setup} We also find that LLMs tend to rely on parametric knowledge 
when it is present, 
and can fail when narratives are inconsistent with the LLM's parametric knowledge. 
To test this,  we elicit the LLM's pre-existing parametric knowledge when generating the event chains. 
We prompt the LLM to 
pick a series of events such that 
each event has some relation to the subsequent event -- either 
the event is \emph{Causal} to the next event 
(e.g., disease causes shorter lives)
or the event is \emph{Anti-Causal} (e.g., disease causes longer lives). 
 For example, we might know that node $1$ is \emph{Anti-Causal} to node $2$ from parametric knowledge. Thus, when we make the causal ground truth graph $1 \rightarrow 3 \rightarrow$ 2  (this disagrees with parametric knowledge), create a narrative from it, and then ask the LLM if node $1$ causes $2$ based on the narrative: it should say yes based on the narrative even though that disagrees with its parametric knowledge. After the ground truth graph is created, we generate the narrative in the \emph{Forward} topological orientation to avoid confounding failure modes. The full process (along with illustration) explaining how the parametric and causal graphs are created is in Appendix \ref{sec:param_appd}. As a textual example, assume that we know a parametric anti-causal link exists from \emph{stressful job} to \emph{increased happiness}, and from \emph{lack of sleep} to \emph{improved cognitive function}. We can then construct the causal chain \emph{Stressful Job} → \emph{Lack of Sleep} → \emph{Increased Happiness} → \emph{Improved Cognitive Function}. From this causal chain, we create the narrative:
\begin{quote}
    The constant demands of a \emph{stressful job} led to her experiencing chronic \emph{lack of sleep}. Surprisingly, she found that the \emph{lack of sleep} heightened her sense of euphoria, making her unusually cheerful at work. \emph{Increased happiness} from this unexpected cheerfulness seemed to improve her \emph{cognitive function}.
\end{quote}
  If the LLM is asked if a stressful job leads to increased happiness, the parametric knowledge shortcut indicates the answer should be no -- however, the shortcut fails as the narrative indicates that a (indirect) causal link does exist.

\paragraph{Models Exploit Parametric Knowledge} We find that, in synthetic experiments, 
the LLM finds the correct causal relation 
generally only when that relation agrees
with its parametric knowledge. This is exemplified in the plot in Figure \ref{fig:param} (left) where we see good performance on narratives that agree with parametric knowledge (\emph{Causal} parametric knowledge) and poor performance on narratives that disagree with parametric knowledge (\emph{Anti-Causal} parametric knowledge). We also notice an interesting phenomenon for the \emph{Anti-Causal} parametric case where using just the extracted graph provides massive improvements over any prompting strategy that involves using the narrative to directly answer questions. This strategy is comparable in performance to when the parametric knowledge is \emph{Causal}.  It seems that the narrative may only serve to distract the LLM when parametric knowledge disagrees with the narrative. 

\subsection{Impact of Narrative Complexity}
In the previous sections, we identified two shortcuts which models exploit in causal reasoning tasks. Here, we test the influence of narrative complexity on these failure modes. We examine two measures of complexity: (a) the narrative length and (b) the presence of complex graph structures.  
\paragraph{Narrative Length} In conditions where the LLM exhibits failure modes (\emph{Reverse} and \emph{Anti-Causal} orientations), the performance also tends to decay as the size of the narrative and the number of events in the narrative increases. As we can see in Figures \ref{fig:forward} and \ref{fig:param} (Left), it seems that the longer the narrative is, the more the LLM relies on shortcuts instead of performing reasoning. However, the extracted graph $G'$ can often maintain a consistently high level of accuracy across narrative sizes even for cases when a failure mode would normally be exhibited.

\label{sec:complexity}
\paragraph{Causal Graph Complexity} As the bulk of our work has focused on detecting the simplest failure modes possible, we studied narratives with an underlying chain graph structure. However, the presence of more complex causal structures in the narrative could exacerbate the existing failure modes or trigger novel failures. To study this, we create causal graphs utilizing two common causal structures: \emph{Forks} (one node has a causal relationship to multiple other nodes) and \emph{Colliders} (multiple nodes have a causal relationship to the same node). We generate narratives (the complete algorithm is described in Appendix \ref{sec:complex_algo}) such that each underlying causal graph contains at least one of these structures, and may randomly contain multiple such structures based on the size of the narrative. An example is shown in Figure \ref{fig:17}.

\begin{figure}[h]
\centering
\begin{minipage}{0.45\textwidth}
\centering
\begin{tikzpicture}[thick]

  \node[draw, rectangle] (rain) at (0, 1.2) {Heavy Rainfall};
  \node[draw, rectangle] (power) at (-1.2, 0) {Power Outage};
  \node[draw, rectangle] (flood) at (1.2, 0) {Flooded Streets};
  \node[draw, rectangle] (traffic) at (0, -1.2) {Traffic Jam};

  \draw[->] (rain) -- (power);
  \draw[->] (rain) -- (flood);
  \draw[->] (power) -- (traffic);
  \draw[->] (flood) -- (traffic);

\end{tikzpicture}
\end{minipage}
\hfill
\begin{minipage}{0.5\textwidth}
\small
\textbf{Narrative:} The \emph{heavy rainfall} not only caused a \emph{power outage} in several neighborhoods but also led to \emph{flooded streets}.  The aftermath of the \emph{power outage} (disabling traffic lights) and the \emph{flooded roads} (blocking street access) caused a \emph{traffic jam}. 
\end{minipage}
\caption{Causal graph with story showing a fork (first sentence) and a collider (second sentence).}
\label{fig:17}
\end{figure}

As can be seen in Figure \ref{fig:param} (right side), we find that while the LLM generally performs worse at reasoning about the complex narratives than simple narratives (with underlying chain graphs), the gap is very starkly less than can be seen in the other failure modes. This finding can be supported by \citep{dettki2025largelanguagemodelsreason} which finds that GPT-4o reasons similarly to humans on a single sentence that describes one collider relation. Our work extends their work by using a long-form narrative based on a causal graph with potentially multiple colliders and forks instead of only one collider.  

\section{Experiments with Real World Causal Graphs}\label{sec:real-world-expt}

In this section, we extend our analysis to narratives involving
real-world causal graphs from \emph{CauseNet} \citep{heindorf2020causenet},
a large-scale knowledge graph of (claimed) causal relationships
between real-world concepts.
We perform experiments using the \emph{GPT-4o} \citep{openai2024gpt4technicalreport} and Llama-3.1 8B models for our experiments. We concentrate our analysis on the same factors (positional biases and parametric knowledge consistency) as explored in the semi-synthetic settings.

\subsection{Experimental Setting}
\label{sec:causenetsetup}

The \emph{CauseNet} dataset can be represented as
a collection of $D$ tuples $\{(C_i, E_i, \mathbf{S}_i\}_{i=1}^{D}$,
where $C_i$ denotes the cause (e.g., fatigue),
$E_i$ denotes the effect (e.g., accidents),
and $\mathbf{S}_i$ is a set of sentences 
(extracted from Wikipedia and ClueWeb12 \citep{callan2012lemur})
that entail a causal relationship from $C_i$ to $E_i$.
We retrieve causal chain graphs
$V_1 \rightarrow V_2 \rightarrow \hdots \rightarrow V_N$
of various lengths,
where each causal relation
$V_i \rightarrow V_{i+1}$ is from \emph{CauseNet} and verbalize these chains as narratives in the following ways:

\paragraph{Semi-synthetic narratives.} In this setting, we use real causal graphs from \textit{CauseNet} but synthetically verbalize them via the LLM. In particular, we prompt the LLM to generate sentences for each edge ($V_i \rightarrow V_{i+1}$) in the causal graph, while ensuring the sensibility of the entire narrative. For eammple, the following is a narrative for the chain
\emph{fatigue} $\rightarrow$ \emph{accidents} $\rightarrow$ \emph{injury}:
\begin{quote}
\emph{Fatigue} can cloud judgment and slow reaction times, leading to an increase in \emph{accidents} on the road.
As a result, these \emph{accidents} often lead to serious \emph{injury} for those involved, highlighting the dangerous consequences of driving while fatigued.
\end{quote}

\paragraph{Real-world narratives.}
For the real-world narratives, 
the sentence for each edge is chosen from the \emph{CauseNet}
dataset.
To ensure that the narrative as a whole remains coherent,
we prompt the LLM to ensure that
the sentences for every pair of adjacent edges
logically follow each other.
For example, the following is the narrative for the causal chain
\emph{fatigue} $\rightarrow$ \emph{accidents} $\rightarrow$ \emph{injury}:
\begin{quote}
Workers work long hours in mines and factories where \emph{fatigue} and a lack of concentration can easily cause \emph{accidents}.
These \emph{accidents} are the leading cause of \emph{injury} in this country for people ages 1-34.
\end{quote}
Additional examples of semi-synthetic and real-world
narratives are presented in Appendix~\ref{sec:apdx-real-world-examples}
(the entire set of narratives used for our experiments 
is available in the linked code).

\paragraph{Prompting Strategies} For simplicity, we limit the prompting techniques used to 
 (see Appendix~\ref{sec:apdx-real-world-prompt-causal-reasoning}
for the prompt templates): \textbf{Standard QA Prompting}, \textbf{Chain-of-Thought} and \textbf{Explicit Causal Graph Extraction}. We evaluate the accuracy for each pair of nodes $(V_i, V_j)$
for the three prompting strategies on the semi-synthetic and
real-world narratives.

\subsection{Impact of Event Ordering and Chain Length}

As described in the previous section,
we verbalize each causal chain graph 
$V_1 \rightarrow V_2 \rightarrow \hdots \rightarrow V_N$
from \emph{CauseNet}
into a narrative in the forward and
reverse topological order.
\begin{figure*}
\centering
\begin{subfigure}[b]{0.45\textwidth}
\centering
\includegraphics[scale=0.3]{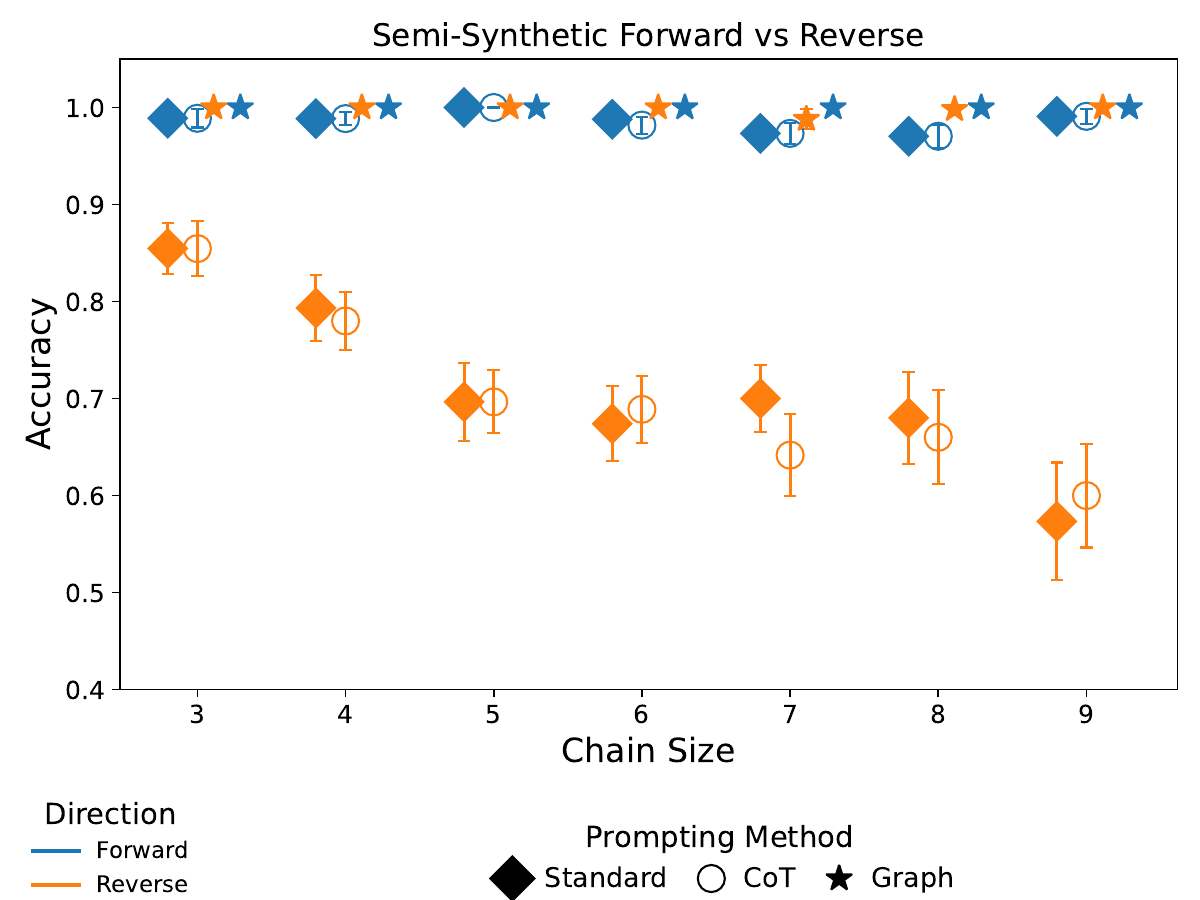}
\label{fig:semi-synth-average}
\end{subfigure}
\hspace{10px}
\begin{subfigure}[b]{0.45\textwidth}
\centering
\includegraphics[scale=0.3]{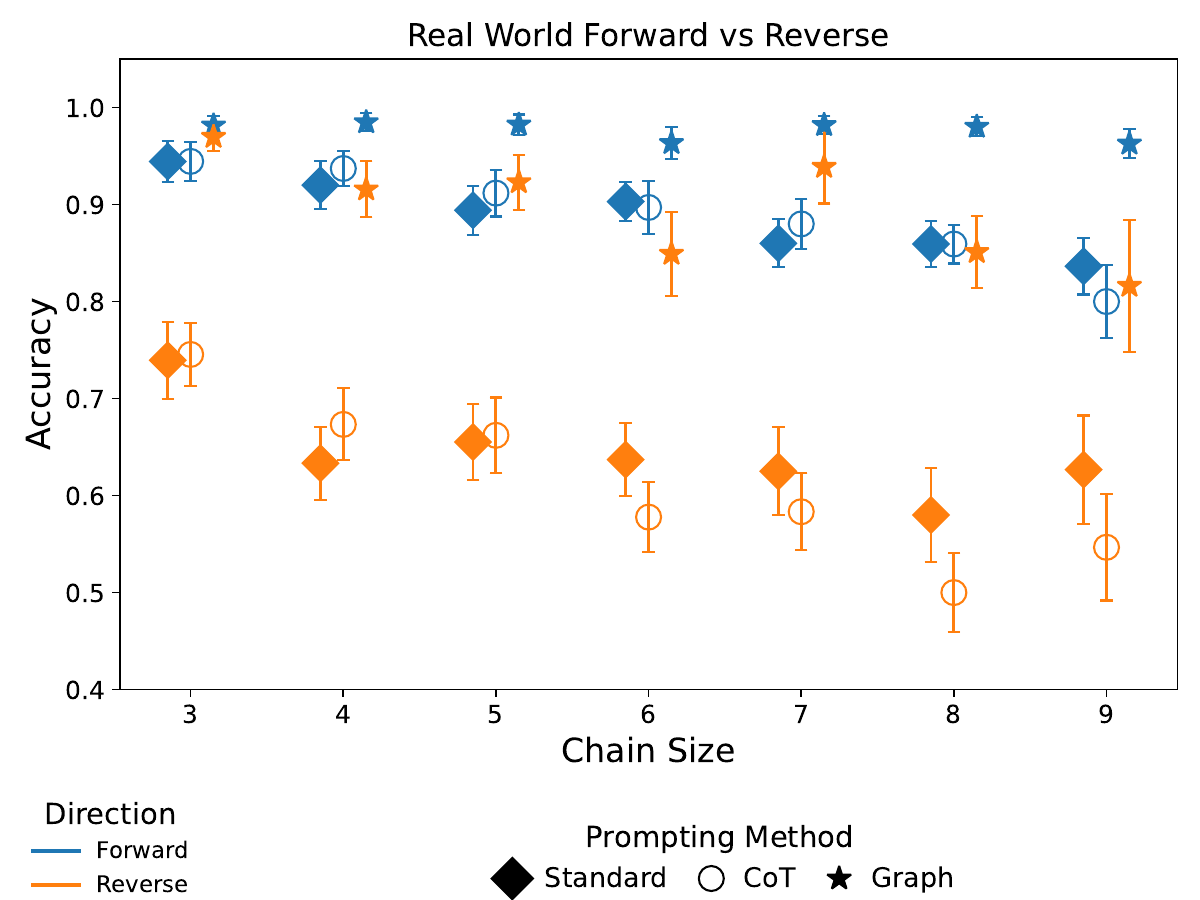}
\label{fig:real-world-average}
\end{subfigure}
\caption{The accuracy of various prompting strategies (error bars denote 95\% CIs).
We observe that the accuracy is lower in the
reverse direction 
(and tends to decay as the chains get longer).
}
\label{fig:average-results1}
\end{figure*}
In both the semi-synthetic (Fig.~\ref{fig:average-results1} left)
and real-world narratives (Fig.~\ref{fig:average-results1} right),
the \emph{Forward Graph} strategy performs the best,
with its accuracy remaining stable even 
as the chain length increases.
We observe that \emph{Forward Standard and CoT}
outperforms \emph{Reverse Standard and CoT},
with the \emph{Reverse} accuracy
declining substantially as the chain size gets large. We also see that in this regime, extracting the causal graph makes inference in the \emph{Reverse} orientation competitive with inference in \emph{Forward}.

\subsection{Effect of Parametric Knowledge Consistency}

\paragraph{Experiment Setup} Next, we analyze the extent to which the
LLM relies on its parametric knowledge to answer causal reasoning
queries as opposed to the causal structure
expressed in the narrative.
For every pair of nodes $(V_i, V_j)$ in the chain graphs,
we elicit the parametric knowledge of the LLM 
by asking the LLM whether a causal effect between the two nodes would be atypical (see Appendix \ref{sec:eliciting} for the exact prompts utilized). Through these prompts, we identify cause and effect chains which contradict the model's parametric knowledge. For example, in a chain graph from our dataset,
there is a  path from 
\emph{streambank erosion} to \emph{higher prices},
but this contradicts the LLM's parametric knowledge
since this causal effect may not typically exist in
the real-world. In total,  we find that roughly 5 percent of the relations in CauseNet violate the LLM's pretraining knowledge. We sampled narratives from CauseNet until we got 100 (of chain sizes between 3 and 9) narratives that contain relations that violate the LLM's pre-training knowledge and 100 that are consistent. These narratives are constructed in the \emph{Forward} topological ordering to avoid confounding failure modes.


\paragraph{LLM Performance Suffers on Atypical Causal Relations}We evaluate the three prompting strategies 
separately on the subsets of cause-and-effect pairs
that are in agreement and in conflict with the
parametric knowledge (see Table~\ref{table:real-world-parametric-conflicts}).
We observe that when there is no conflict
(i.e., the parametric knowledge agrees with the 
causality expressed in the narrative),
the accuracies with and without CoT
are greater than $90\%$.
However, when the parametric knowledge conflicts with
the narrative's causality,
the accuracy is
significantly lower,
even with CoT.
This suggests that when asked to reason about
cause and effect in a narrative, 
the LLM seems to rely heavily on
its parametric knowledge and 
is unable to grasp the specific causal chains 
expressed in the narrative itself 
(despite the causal chains as a whole being realistic). 

\paragraph{Explicit Causal Graph Extraction Avoids Shortcuts} Interestingly, when using extracted graph 
for performing causal reasoning,
the performance is very high, both with and without conflicts.
This is likely because when asked to extract the graph
from the narrative,
the LLM pays more attention to the entire narrative 
as opposed to when directly queried on a cause-and-effect pair 
(where the LLM defaults to its parametric knowledge).
These results show that even when
the LLM constructs a reasonably 
good causal chain graph,
the LLM does not leverage this graph
when queried directly about the causal effects 
in the narrative (even with CoT),
further highlighting the advantage of
extracting the causal graph directly. 
\begin{table*}
\centering
\begin{tabular}{@{}lccc@{}}
                 & \textbf{Standard} & \textbf{CoT} & \textbf{Graph}  \\
\toprule
\multicolumn{4}{l}{\textbf{Semi-synthetic}}                                                                      \\
\midrule
Without Conflict &  99.8                 &    99.6           & 99.9                                  \\
With Conflict    & 67.2                              & 73.1                 & 98.7                  \\
\midrule
\multicolumn{4}{l}{\textbf{Real-world}}                                                                          \\
\midrule
Without Conflict & 90.9                & 89.2                & 97.9                                    \\
With Conflict    & 52.1               & 57.6                & 93.2                    \\    
\bottomrule
\end{tabular}
\caption{The average accuracy across different narratives with the three prompting strategies partitioned by whether the cause-effect pairs
conflict with the LLM's parametric knowledge (we omit the 95\% CIs as they are smaller than $0.3$). 
}

\label{table:real-world-parametric-conflicts}
\end{table*}

\begin{figure*}

    \begin{subfigure}[b]{0.45\textwidth}
\centering
\includegraphics[scale=0.3]{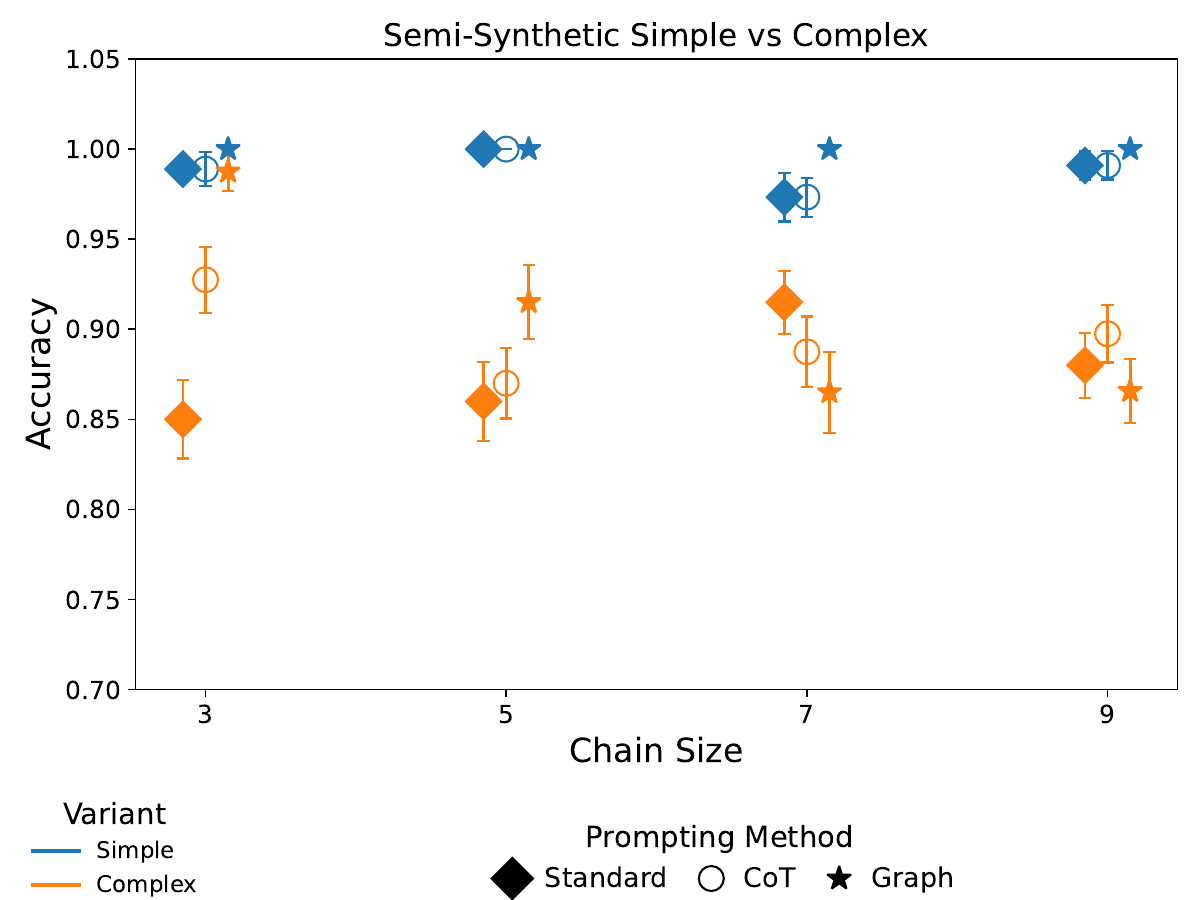}
\end{subfigure}
\hspace{10px}
\begin{subfigure}[b]{0.45\textwidth}
\centering
\includegraphics[scale=0.3]{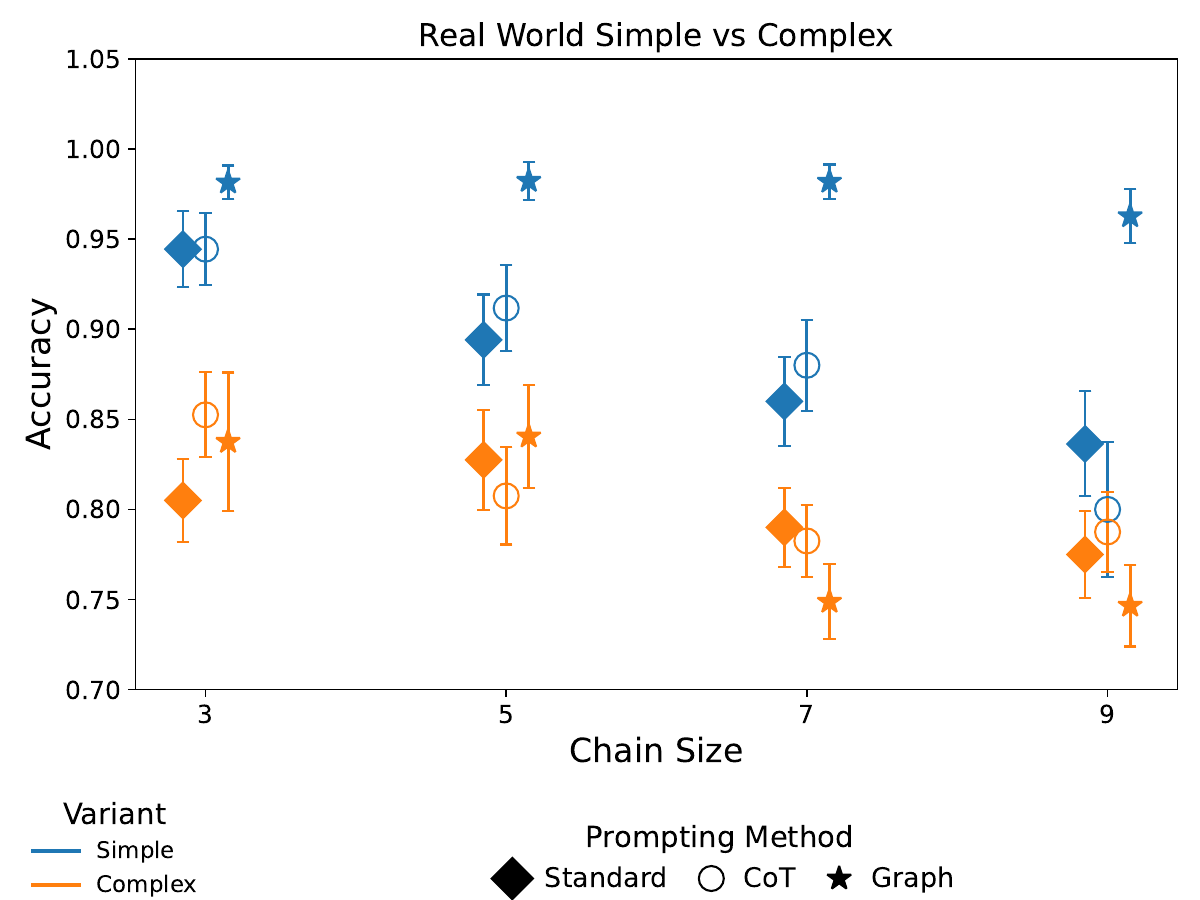}
\end{subfigure}

    \caption{ GPT-4o accuracy on narratives generated from Complex graphs as opposed to Simple chain graphs for semi-synthetic narratives (left) and real-world narratives (right).  95 $\%$ CI is shown. 
    \label{fig:param7}}

\end{figure*}


\subsection{Narrative Complexity}

We can see from Figure \ref{fig:average-results1} that LLM performance degrades with narrative length, especially when a failure mode is present. We furthermore experimented with complex narratives with causal graphs containing forks and colliders (full graph and narrative creation algorithm in Appendix \ref{sec:complex_algo1}). We can see in Figure \ref{fig:param7}, that in both the semi-synthetic and real-world settings that complex narratives (with colliders and forks) perform worse than simple narratives that have a causal chain graph as the ground truth. This gap ,while clear and noticeable, isn't as stark as failure from parametric knowledge conflict (Table \ref{table:real-world-parametric-conflicts})  or topological ordering (Figure \ref{fig:average-results1}). We do furthermore note that this is one area where extracting an explicit causal graph does not seem to significantly improve performance.

\section{Discussion}
Our work takes initial strides
towards examining  
the success and failure of LLMs to reason causally on narratives that express causal events. 
We focus on two questions of key importance 
in causality: 
(1) Does one event cause another? 
(2) Can the LLM extract the causal graph from the narrative.
We find three significant failure modes of LLM reasoning by conducting experiments in carefully controlled synthetic, semi-synthetic and real-world settings: Firstly, we find that LLMs rely heavily on \textbf{topological ordering}, performing well when the ordering of events in the narratives matches that of the ordering of the underlying causal graph. Secondly, we find that LLMs rely on their \textbf{parametric knowledge} as a shortcut to infer causal relations. Finally, we examine the role of \textbf{causal structure complexity}, finding that LLM accuracy degrades as the narrative length increases.  Furthermore, LLMs perform slightly worse on reasoning when narratives contain structures such as colliders and forks.
Beyond these failure modes, we show that more reliable causal reasoning can be elicited by prompting the LLM to explicitly identify the causal graph. 



\subsection{Limitations and Future Works}
One limitation of our work is that 
there are other forms of causal reasoning that 
we did not test for in the narratives. 
This motivates many potential directions for future work. 
For example, it could be interesting to ask the LLM to reason about counterfactual cases. Our analysis also has implications for algorithmic interventions to improve causal reasoning. The failure modes we identify in this paper could inform the design of targeted synthetic tasks to use in finetuning for improved causal reasoning. Additionally, our findings on the benefits of extracting a causal graph can inform prompt engineering efforts to elicit reliable causal reasoning from language models. We believe investigating both directions represents an exciting direction for future work.

\newpage
\bibliography{example_paper}
\bibliographystyle{unsrtnat}

\newpage
\appendix
\section*{Technical Appendices and Supplementary Material}
\section{Synthetic Data Experiments}\label{sec:apdx-synthetic}

\subsection{Selected Synthetic Prompts}
 We use an LLM to generate the events \( \emph{E} \).
From the events, we create a ground truth causal graph \( \emph{G} \) which is used to structure and inform the narrative sequence and causality. \( \emph{N} \) is the corresponding narrative created by the LLM from \( \emph{G} \).  To evaluate the LLM's performance, we extract a causal graph, \( \emph{G'} \), from the narrative \( \emph{N} \) as produced by the LLM, and compare it with the ground truth causal graph \( \emph{G} \). In this context, \( \emph{n} \) refers to the number of events to generate, while \( \emph{A} \) and \( \emph{B} \) represent pairs of events queried for causal relationships. The task then becomes assessing whether event \( \emph{A} \) causes event \( \emph{B} \) . All prompts, data processing steps, and results are included in the attached code.
\subsubsection{Topological Experiment - Generating Random Events (\emph{E})}
``generate \emph{n} random distinct events"
\subsubsection{Parametric Experiment  -Generating a Pair of Causal Events (\emph{E})}
``generate a pair of events that cause each other. generate an event that causes another event, for example Cancer $\rightarrow$ Death or Obesity $\rightarrow$ Bad Heart Health. Make sure the event generated is not already in $E$ "
\\
This is repeated as many times as is necessary
\subsubsection{Parametric Experiment - Generating a Pair of Anti-Causal Events (\emph{E}) }
``generate a pair of events that are anticausal (an event causing the opposite of the normal effect),  for example the first event could be cancer and the second event could be a longer life 
because in reality, cancer causes a shorter life. Make sure the events generated are not already in $E$."
\\
This is repeated as many times as is necessary
\subsubsection{Forward Topological Narrative (\emph{N})}
``Output a short narrative (use one sentence) that expresses the causal link [{E1} $\rightarrow$ {E2}]. By causal link, we mean that the sentence should convey that  
                            {E1} directly caused {E2}. In other words, it should be clear from the narrative that {E2} would not have happened had {E1} not happened. Ensure that the words [{E1}, {E2}]
                            are present in the new sentence and {E1} appears before {E2}. Only output the new sentence."
\\
Repeat for all causal/anti-causal links
\subsubsection{Reverse Topological Narrative (\emph{N})}
``Output a short narrative (use one sentence) that expresses the causal link [{E1} $\rightarrow$ {E2}]. By causal link, we mean that the sentence should convey that  
                            {E1} directly caused {E2}. In other words, it should be clear from the narrative that {E2} would not have happened had {E1} not happened. Ensure that the words [{E1}, {E2}]
                            are present in the new sentence and {E2} appears before {E1}. Only output the new sentence."
                            \\
Repeat for all causal/anti-causal links
\subsubsection{Standard Prompt}
``Use this narrative \emph{N} as context. Did \emph{A}  cause \emph{B}? Output your answer with $<answer>Yes/No</answer>$. The cause can be direct or indirect."
\subsubsection{Chain of Thought Prompt}
``Use this narrative \emph{N} as context. Did \emph{A}  cause \emph{B}? Do step by step reasoning. Then output your answer with $<answer>Yes/No</answer>$. The cause can be direct or indirect."
\subsubsection{In-Context Prompt}
``Use this narrative \emph{N} as context. Did \emph{A}  cause \emph{B}? Output your answer with $<answer>Yes/No</answer>$. The cause can be direct or indirect. 
 An example narrative would be: Rains leads to plants growing. This then causes increased oxygen in the atmosphere. A potential question would be does rain cause increased oxygen in the atmosphere? The answer would be Yes.
 Another example narrative would be: Increased oxygen in the atmosphere is because of plants growing. Plants grow because rain provides them essential nutrients. A potential question would be does rain cause increased oxygen in the atmosphere? The answer would be Yes.
 Another example narrative would be: Rain leads plants to grow. Plants growing causes less oxygen in the atmosphere. A potential question would be does rain cause less oxygen in the atmosphere? The answer would be Yes."

\subsubsection{Narrative + Graph Prompt}
``Use this narrative \emph{N} and this causal ordering \emph{G'} ((such that each item is a cause of every item after it, for example the first list item is a cause of the third, fourth, fifth items etc)) as context. Did \emph{A}  cause \emph{B}? Output your answer with $<answer>Yes/No</answer>$. The cause can be direct or indirect."
\subsection{Parametric Graph Experiment}
\label{sec:param_appd}
\begin{figure*}[h!]

\centering
\includegraphics[scale=0.35]{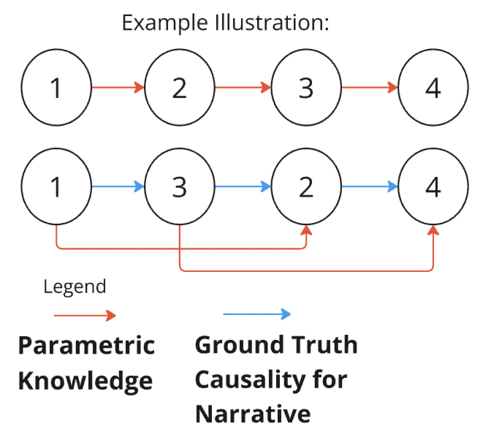}

    \caption{ Example illustration (right)  is of how $G$, the ground truth causality, is set up. }
    \label{fig:param188}

\end{figure*}

Let's call the graph of parametric knowledge $P$. 
We then take the odd indexed events (1st, 3rd etc) from $P$ and place them in the first half of the causal ground truth graph $G$ and the even indexed events (2nd, 4th etc) from $P$ in the second half of $G$. This process is shown in Figure \ref{fig:param188}.

\subsection{Complex Graph Creation}

\label{sec:complex_algo}
To generate a ground-truth causal graph \emph{G} with rich structure (colliders, forks, and a spanning chain), for each choice of size \(n\) we perform the following algorithm:

\begin{enumerate}
  \item \textbf{Node sampling.}  Draw \(n\) distinct events
    \[
      \{E_1, E_2, \dots, E_n\} \;\subset\; \mathcal{E}
    \]
    uniformly at random without replacement.

    \item \textbf{Determine motif counts.}  (for \(n\ge4\))
    \[
      k_{\max} \;=\;\bigl\lfloor n/2\bigr\rfloor,\quad
      k_{\mathrm{tot}}\sim \mathrm{Uniform}\bigl(2,\,k_{\max}\bigr),
    \]
    \[
      k_{\mathrm{col}}\sim \mathrm{Uniform}\bigl(1,\,k_{\mathrm{tot}}-1\bigr),
      \qquad
      k_{\mathrm{fork}} = k_{\mathrm{tot}} - k_{\mathrm{col}}.
    \]

  \item \textbf{Collider creation.}  Repeat \(k_{\mathrm{col}}\) times:
    \begin{enumerate}
      \item Select two distinct “parent” nodes \(p_1,p_2\) from those not yet used in any motif.
      \item Select a “child” node \(c\) that is neither \(p_1\) nor \(p_2\) and not yet used as a child.
      \item Add edges
        \[
          p_1 \;\to\; c
          \quad\text{and}\quad
          p_2 \;\to\; c\,,
        \]
        thereby forming a collider at \(c\).
    \end{enumerate}

  \item \textbf{Fork creation.}  Repeat \(k_{\mathrm{fork}}\) times:
    \begin{enumerate}
      \item Select a “parent” node \(p\) from those not yet used.
      \item Select two distinct “child” nodes \(c_1,c_2\) from the remaining unused nodes.
      \item Add edges
        \[
          p \;\to\; c_1
          \quad\text{and}\quad
          p \;\to\; c_2\,,
        \]
        forming a fork with shared parent \(p\).
    \end{enumerate}

  \item \textbf{Chain-connect remaining nodes.}  Let
    \(\mathcal{R}\) be the set of nodes not yet involved in any collider or fork.
    \begin{enumerate}
      \item Order \(\mathcal{R} = \{r_1,\dots,r_m\}\) arbitrarily, then add chain edges
        \[
          r_1 \to r_2,\;\; r_2\to r_3,\;\;\dots,\;\; r_{m-1}\to r_m.
        \]
      \item To ensure the entire graph is connected, choose one node
        \(u\) from among the previously used nodes (if any) and add
        \[
          u \;\to\; r_1\,.
        \]
    \end{enumerate}
\end{enumerate}

\section{Real-world Causal Graphs}

\subsection{Prompt templates for narrative generation}\label{sec:apdx-real-world-prompt-narrative-gen}

Recall that we have a ground truth causal chain graph of
the form $V_1 \rightarrow V_2 \rightarrow \hdots \rightarrow V_N$
from \emph{CauseNet} that we need to verbalize into a coherent narrative.
For the semi-synthetic narratives, we use the LLM (GPT-4o)
to do so one edge at a time, 
while ensuring that the newly verbalized edge logically
follows the previous one.
The following is the prompt template for generating the
narratives in the topological order of the graph:
\begin{quote}
Output a short narrative (use one or two sentences) 
that expresses the causal link 
[$V_i \rightarrow V_{i+1}$] and logically follows this narrative:

\{ Narrative for the previous edge $V_{i-1} \rightarrow V_{i}$\}.

Ensure that the combined sentences convey the causal chain 
[$V_{i-1} \rightarrow V_{i} \rightarrow V_{i+1}$] and 
that the words [$V_i$, $V_{i+1}$] are present.
Only output the newly generated narrative. 
\end{quote}
Similarly, we generate narratives in the reverse topological
order of the graph by verbalizing edges in the reverse direction
with the following prompt template:
\begin{quote}
Output a short narrative (use one or two sentences) 
that expresses the causal link 
[$V_i \rightarrow V_{i+1}$] and logically follows this narrative:

\{ Narrative for the previous edge $V_{i+1} \rightarrow V_{i+2}$\}.

Ensure that the combined sentences convey the causal chain 
[$V_{i} \rightarrow V_{i+1} \rightarrow V_{i+2}$] and 
that the words [$V_i$, $V_{i+1}$] are present.
Only output the newly generated narrative.
\end{quote}

For generating real-world narratives, for each edge $V_i \rightarrow V_j$,
we use the set of sentences from \emph{CauseNet}.
Each edge in \emph{CauseNet} is linked 
to multiple sentences from various sources. 
Picking a sentence for each edge at random and concatenating
them does not always lead to sensible narratives.
To improve the quality of narratives, we use the following prompt
to concatenate sentences for adjacent edges:
\begin{quote}
Consider the following sentences.

\{ Sentence for edge $V_{i} \rightarrow V_{i+1}$ \}. \{ Sentence for edge $V_{i+1} \rightarrow V_{i+2}$ \}.

Do the sentences logically follow each other and express the causal chain [$V_{i} \rightarrow V_{i+1} \rightarrow V_{i+2}$]? Answer with Yes or No.
\end{quote}
For verbalizing narratives in the topological order,
for a given graph $V_1 \rightarrow V_2 \rightarrow \hdots \rightarrow V_N$,
we only use sentences such that the above prompt returns \emph{Yes}
for every pair of adjacent edges $V_{i} \rightarrow V_{i+1} \rightarrow V_{i+2}$.
This ensures that the narrative as a whole remains coherent
and conveys the entire causal chain graph.
We use a similar prompting strategy to verbalize narratives
in the reverse topological order.

\subsection{Eliciting Parametric Knowledge}
\label{sec:eliciting}
We ask the LLM ``Does $V_i$ typically have a causal (indirect or direct) effect on $V_j$?'' and ``Would it be atypical if $V_i$ had a (indirect or direct) causal effect on $V_j$?''. If the LLM answers ``No'' and ``Yes'' to those respective questions, we would consider a causal relationship between $V_i$ and $V_j$ to contradict the LLM's prior knowledge that it learned from its pretraining corpora. 
\subsection{Semi-Synthetic and Real-World Complex Graph Algorithm}
\label{sec:complex_algo1}
Let \(\mathcal{M}=\{(u,v)\}\) be the set of real‐world causal edges from CauseNet.  For each target size \(n\in\{3,\dots,9\}\), we:

\begin{enumerate}
  \item \textbf{Load CauseNet.}  
    \[
      \mathcal{M} \;=\;\bigl\{(u,v)\mid u\!\to\!v\text{ in CauseNet}\bigr\}.
    \]

  \item \textbf{Extract collider and fork motifs.}
    \begin{align*}
      \mathsf{Colliders} &= \{(p_1,p_2,c)\mid (p_1,c)\in\mathcal{M},\,(p_2,c)\in\mathcal{M},\,p_1\neq p_2\}, \\
      \mathsf{Forks}     &= \{(r,c_1,c_2)\mid (r,c_1)\in\mathcal{M},\,(r,c_2)\in\mathcal{M},\,c_1\neq c_2\}.
    \end{align*}

  \item \textbf{Determine motif counts.}  
    \[
      \text{If }n=3,\quad
      (k_{\mathrm{col}},k_{\mathrm{fork}})
      =\begin{cases}
        (1,0)\quad&\text{w.p.\ }0.5,\\
        (0,1)\quad&\text{w.p.\ }0.5.
      \end{cases}
    \]
   (for \(n\ge4\))
    \[
      k_{\max} \;=\;\bigl\lfloor n/2\bigr\rfloor,\quad
      k_{\mathrm{tot}}\sim \mathrm{Uniform}\bigl(2,\,k_{\max}\bigr),
    \]
    \[
      k_{\mathrm{col}}\sim \mathrm{Uniform}\bigl(1,\,k_{\mathrm{tot}}-1\bigr),
      \qquad
      k_{\mathrm{fork}} = k_{\mathrm{tot}} - k_{\mathrm{col}}.
    \]

  \item \textbf{Select motifs.}  
    \begin{itemize}
      \item Sample \(k_{\mathrm{col}}\) distinct triples from \(\mathsf{Colliders}\).
      \item Sample \(k_{\mathrm{fork}}\) distinct triples from \(\mathsf{Forks}\).
    \end{itemize}
    Let \(S\) be the union of all nodes appearing in these sampled triples.

  \item \textbf{Pad or trim to size \(n\).}  
    \begin{itemize}
      \item If \(|S|>n\), uniformly subsample \(n\) nodes from \(S\).
      \item If \(|S|<n\), add random “seed” nodes (not already in \(S\)) until \(|S|=n\).
    \end{itemize}

  \item \textbf{Build ground‐truth edges \(\mathcal{G}\subseteq S\times S\).}  
    \begin{enumerate}
      \item \emph{Colliders:} for each \((p_1,p_2,c)\) chosen, add
        \(p_1\to c\) and \(p_2\to c\).
      \item \emph{Forks:}    for each \((r,c_1,c_2)\), add
        \(r\to c_1\) and \(r\to c_2\).
      \item \emph{Chains:}   for any remaining \((u,v)\in S\times S\)
        with \((u,v)\in\mathcal{M}\) and neither \(u\) nor \(v\) used
        in the above, add \(u\to v\) to ensure connectivity.
    \end{enumerate}

  \item \textbf{Narrative generation.}  For each \((u\!\to\!v)\in\mathcal{G}\): \\
    For the semi-synthetic case -  prompt the LLM to generate a sentence linking u to v using the forward topological ordering prompt. \\
    For the real-world case: Find a causal sentence linking u and v in the Cause-Net database

\end{enumerate}

\section{Real-World Complex Graph Creation}

\subsection{Additional examples of the generated narratives}\label{sec:apdx-real-world-examples}

\subsubsection{Semi-synthetic narratives}

Below, we present some examples of 
semi-synthetic narratives in the
forward and reverse directions.

The narrative in the forward direction for the chain
\emph{higher prices} $\rightarrow$ \emph{reduced demand} $\rightarrow$ \emph{lower prices}:
\begin{quote}
As \emph{higher prices} swept through the market, consumers began to tighten their budgets, leading to a noticeable \emph{reduction in demand} for many goods.
As a result of the \emph{reduced demand}, suppliers were forced to \emph{lower prices} in order to attract buyers back to the market.
\end{quote}

The narrative in the reverse order for the causal chain
\emph{bankruptcy} $\rightarrow$ \emph{bad credit} $\rightarrow$ \emph{rejection} $\rightarrow$ \emph{anger}:
\begin{quote}
The sting of rejection ignited a fire within her,
transforming her hurt into a seething anger that demanded to 
be felt.
Her bad credit had led to the rejection she 
never saw coming, and now that sting of rejection ignited a 
fire within her, transforming her hurt into a seething anger 
that demanded to be felt.
Her bankruptcy had left her 
with bad credit, a shadow that loomed over her every 
application, and now that sting of rejection ignited a fire 
within her, transforming her hurt into a seething anger that 
demanded to be felt.
\end{quote}

The narrative in the reverse order for the causal chain
\emph{pollution} $\rightarrow$ \emph{climate change} $\rightarrow$ \emph{extreme weather events} $\rightarrow$ \emph{natural disasters}:
\begin{quote}
As extreme weather events become more frequent and severe, they increasingly lead to devastating natural disasters that disrupt communities and ecosystems alike.
Climate change is driving the rise in extreme weather events, which in turn are causing unprecedented natural disasters that threaten the stability of communities and the health of ecosystems.
Pollution is a major contributor to climate change, which is driving the rise in extreme weather events that threaten the stability of communities and the health of ecosystems.
\end{quote}

\subsubsection{Real-world narratives}

Below, we present some examples of 
real-world narratives in the
forward and reverse directions.

The narrative in the forward direction for the chain
\emph{higher prices} $\rightarrow$ \emph{reduced demand} $\rightarrow$ \emph{lower prices}:
\begin{quote}
\emph{Higher prices} generally lead to reduced demand.
\emph{Lower prices}, caused by \emph{reduced demand} and increased competition for soybeans and corn, largely contributed to the overall bulk export decline.
\end{quote}

The narrative in the reverse order for the causal chain
\emph{bankruptcy} $\rightarrow$ \emph{bad credit} $\rightarrow$ \emph{rejection} $\rightarrow$ \emph{anger}:
\begin{quote}
Embittered by an abusive upbringing, seething with resentment, irritated by others' failure to fulfill his or her superior sense of entitlement, and fuelled by anger resulting from rejection, the serial bully displays an obsessive, compulsive and self-gratifying urge to displace their uncontrolled aggression onto others whilst exhibiting an apparent lack of insight into their behavior and its effect on people around them.
Bad credit normally leads to rejection but now with bad credit secured loan, you can avail the loan of your choice.
For example, if you are applying for a loan, the lender may reject your application on the basis of bad credit caused by bankruptcy.
\end{quote}

The narrative in the reverse order for the causal chain
\emph{pollution} $\rightarrow$ \emph{climate change} $\rightarrow$ \emph{extreme weather events} $\rightarrow$ \emph{natural disasters}:
\begin{quote}
In addition to forced migrations from rising seas, climate change is also 
increasing extreme weather events causing natural disasters such as cyclonic storms (hurricanes or typhoons), floods and droughts.
This is worsened by extreme weather events caused by climate change.
This landmark bill would jump start the economy by creating millions of new clean energy jobs, increase national security by reducing dependence on foreign oil, and preserve the planet by reducing the pollution that causes climate change.
\end{quote}

\subsection{Prompt templates for assessing causal reasoning}\label{sec:apdx-real-world-prompt-causal-reasoning}

We use the following template for the Direct prompting strategy:
\begin{quote}
Consider the following hypothetical narrative.

\{narrative\}

According to the hypothetical narrative, does \{cause\} have a 
(direct or indirect) causal effect on \{effect\}?
Answer in Yes/No.
\end{quote}

We use the following template for the Chain-of-Though (CoT) prompting strategy:
\begin{quote}
Consider the following hypothetical narrative.

\{narrative\}

According to the hypothetical narrative, does \{cause\} have a 
(direct or indirect) causal effect on \{effect\}?
Think step-by-step and end your answer with \textless{answer}\textgreater Yes/No\textless{/answer}\textgreater.
\end{quote}

We use the following template to extract a chain graph from the narrative:
\begin{quote}
Consider the following hypothetical narrative.

\{narrative\}

According to the hypothetical narrative, construct a causal chain graph using the following nodes: \{ nodes in random order \}.
Ensure that the graph contains all the given nodes and only output a single chain graph of the form \textless{graph}\textgreater node1 $\rightarrow$ node2 $\rightarrow$ node3 \textless{/graph}\textgreater.
Only output the graph between the \textless{graph}\textgreater \textless{/graph}\textgreater tags.
\end{quote}

\subsection{Necessary Compute}
No pretraining was done so no GPUs were needed. We used cloud based API calls to pre-trained models like ChatGPT, Anthropic and Llama. We estimate that for the synthetic portion, our API calls to ChatGPT, Anthropic and LLama took 10 hours each. For the semi-synthetic and real-world portion, we had roughly 10 hours of API calls for ChatGPT and Llama each. So in total, roughly 50 hours of API usage. As the majority of the computational burden fell on cloud based API calls, no significant CPU resources are required either.
\newpage

\section{Additional Results - Synthetic Data}

\subsection{Forward vs Reverse Experiments
Anthropic and LLama}
\begin{figure*}[h!]
    
    \begin{subfigure}[b]{0.45\textwidth}
\centering
\includegraphics[scale=0.32]{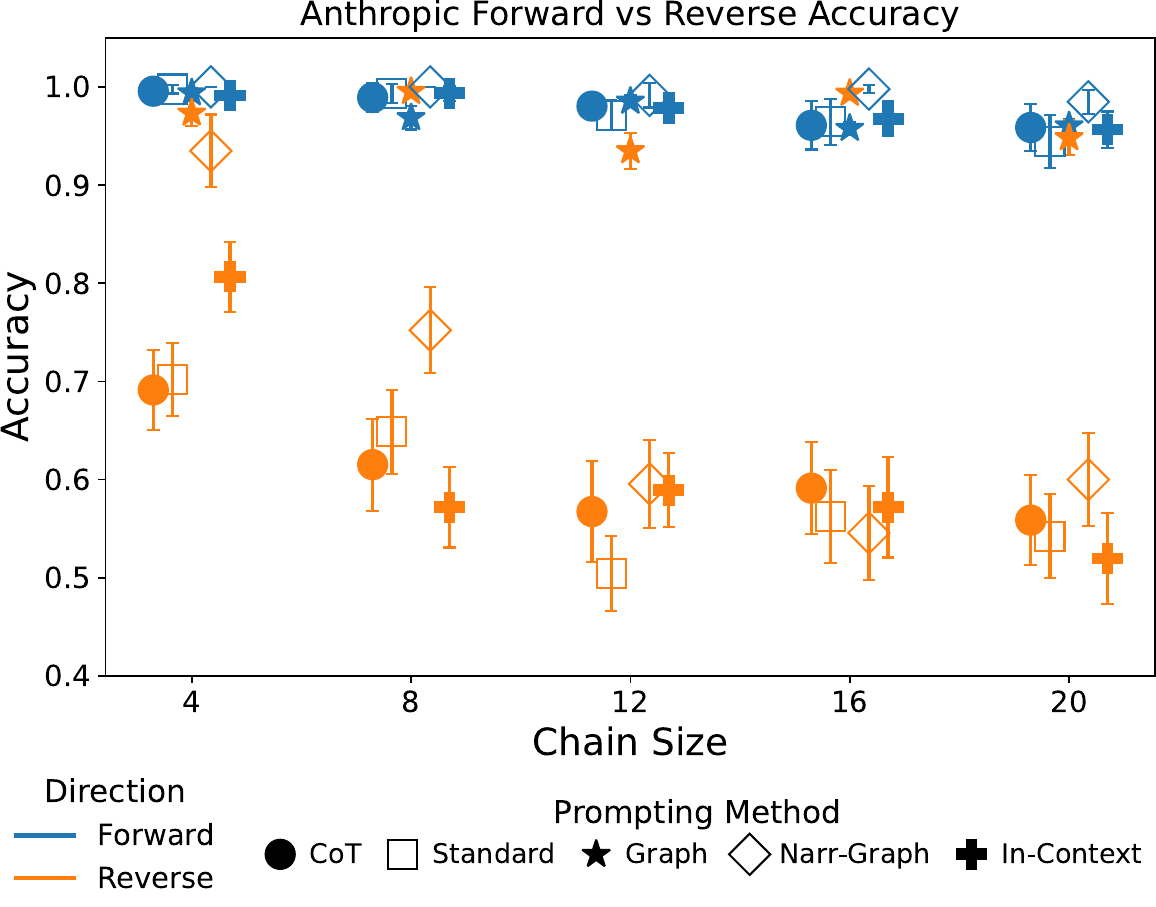}
\subcaption{Anthropic Claude 3.5 Sonnet}
\end{subfigure}
\hspace{10px}
\begin{subfigure}[b]{0.45\textwidth}
\centering
\includegraphics[scale=0.32]{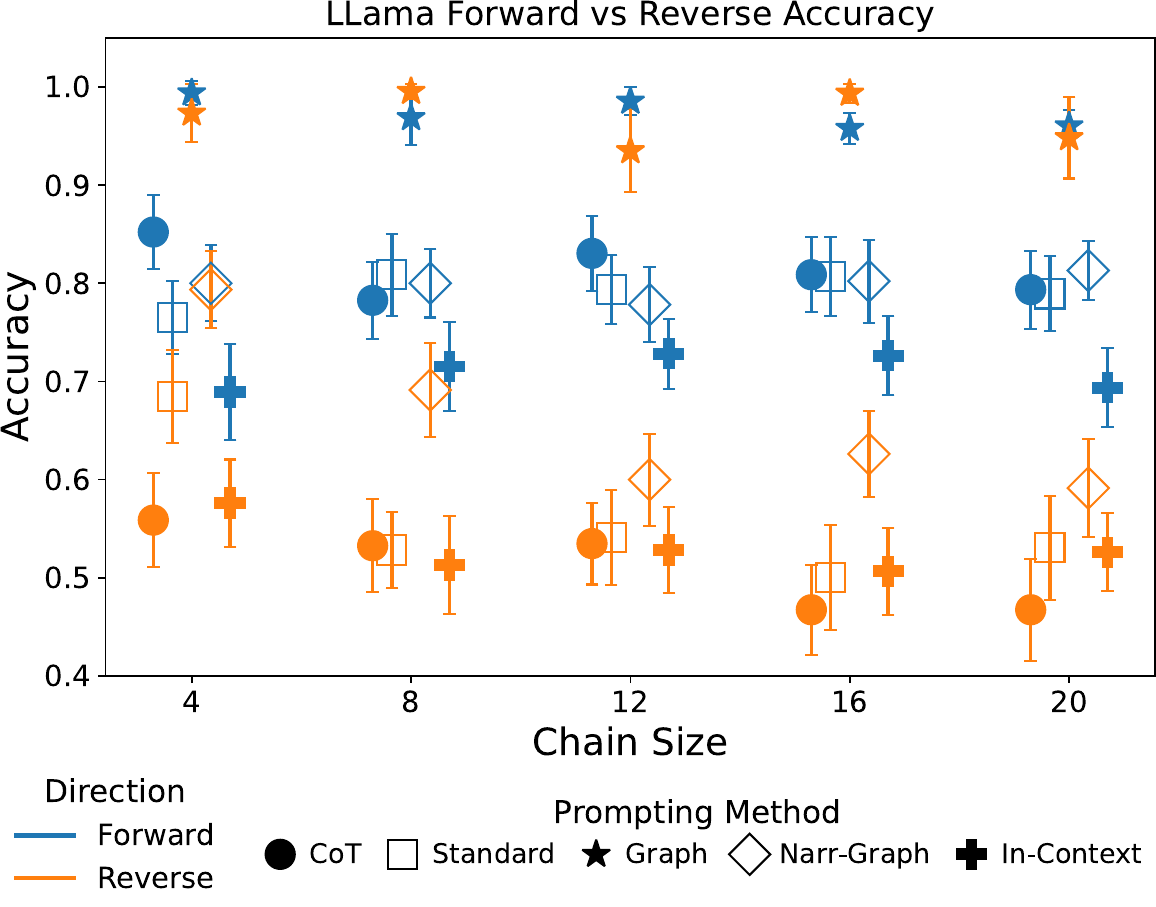}
\subcaption{LLama 3.1 8B}
\end{subfigure}

    \caption{(a) Anthropic Claude 3.5 Sonnet and (b) LLama 3.1 8B Test of the LLM's ability to reason on narratives written in the Forward and Reverse topological orientations. Chain size is the number of nodes in ground truth $G$. The "Graph" prompting method uses only the extracted graph $G'$ to reason,  "Narr-Graph" uses both the narrative and extracted graph, and "Standard, CoT, In-Context" all use only the narrative.
    Accuracy measures LLM answer agreement with $G$. The points in the graph are represented with a slight horizontal stagger around the relevant chain sizes (4,8,12 etc)  for ease of visual understanding. 
    We show a 95$\%$ CI.}
    
\end{figure*}

\subsection{Causal Vs Anti-Causal Experiments Anthropic and LLama}

\begin{figure*}[h!]

    \begin{subfigure}[b]{0.45\textwidth}
\centering
\includegraphics[scale=0.32]{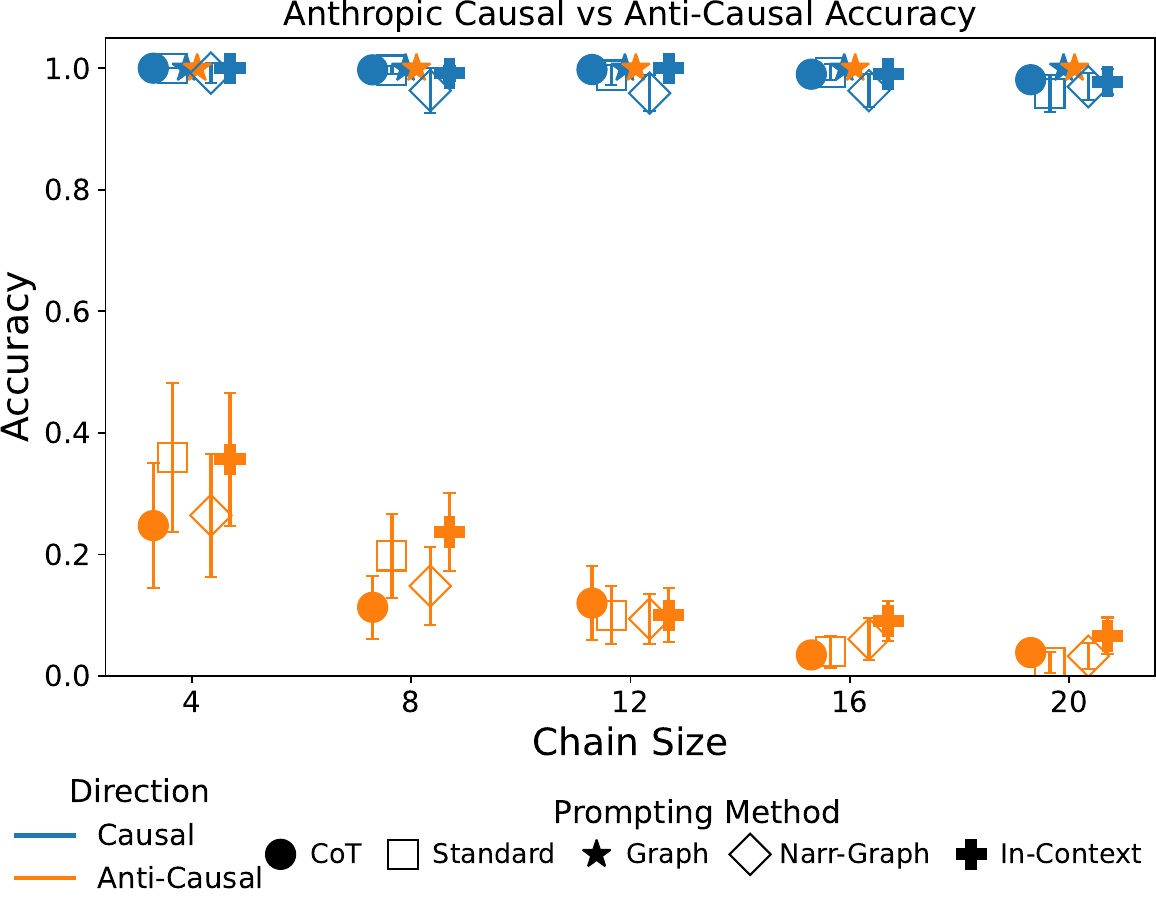}
\subcaption{Anthropic Claude 3.5 Sonnet}
\end{subfigure}
\hspace{10px}
\begin{subfigure}[b]{0.45\textwidth}
\centering
\includegraphics[scale=0.32]{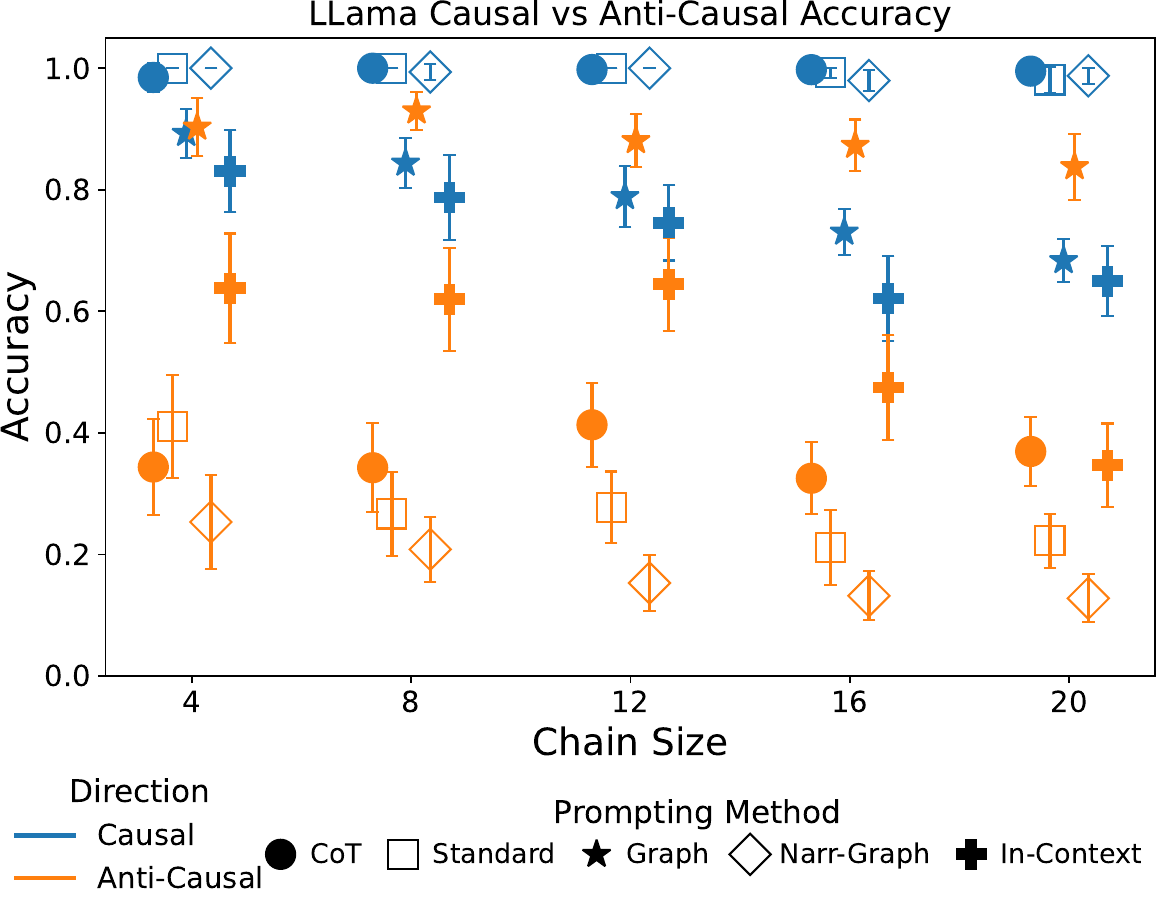}
\subcaption{LLama 3.1 8B}
\end{subfigure}

    \caption{ (a) Anthropic Claude 3.5 Sonnet and (b) LLama 3.1 8B Test of the LLM's ability to reason on narratives that agree with parametric knowledge (Causal) and disagree with parametric knowledge (Anti-Causal). 95 $\%$ CI is shown. }
\end{figure*}

\newpage
\subsection{Complex vs Simple Graphs Anthropic and LLama}
\begin{figure*}[h!]

    \begin{subfigure}[b]{0.45\textwidth}
\centering
\includegraphics[scale=0.32]{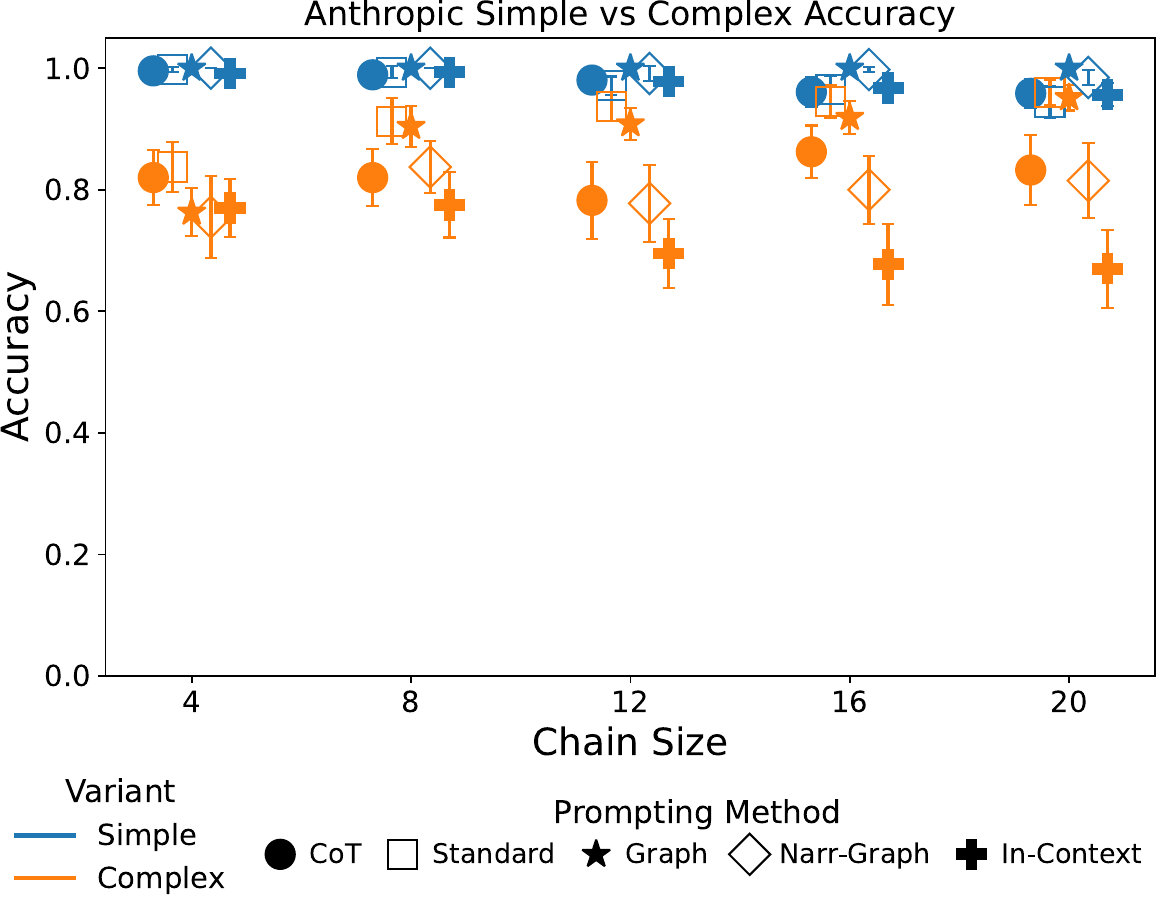}
\subcaption{Anthropic Claude 3.5 Sonnet}
\end{subfigure}
\hspace{10px}
\begin{subfigure}[b]{0.45\textwidth}
\centering
\includegraphics[scale=0.32]{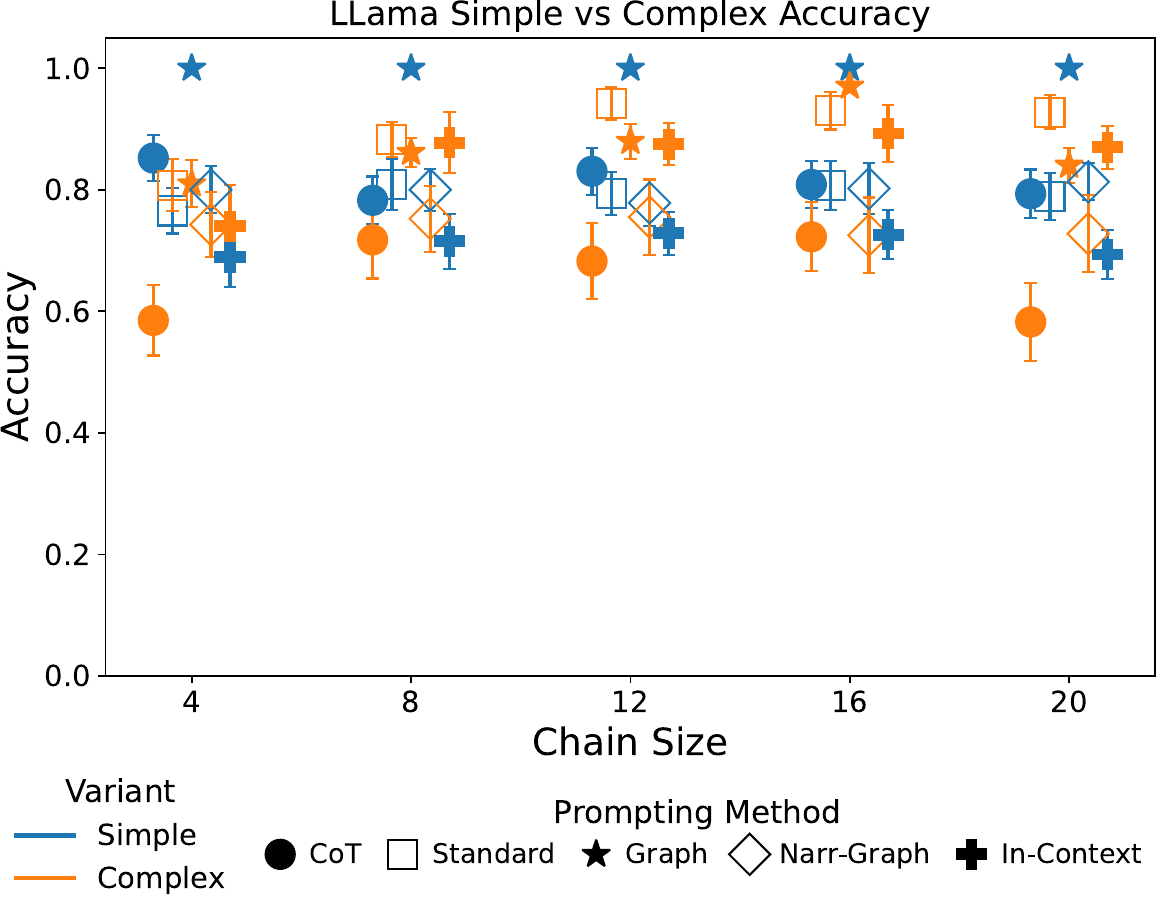}
\subcaption{LLama 3.1 8B}
\end{subfigure}

    \caption{ (a) Anthropic Claude 3.5 Sonnet and (b) LLama 3.1 8B Test of the LLM's ability to reason on narratives generated from Complex graphs as opposed to Simple chain graphs. 95 $\%$ CI is shown. }
\end{figure*}

\newpage

\section{Additional results - Semi-Synthetic and Real World Data}\label{sec:apdx-real-world-additional-results}
\subsection{Forward vs Reverse LLama}
\begin{figure*}[h!]
\centering
\begin{subfigure}[b]{0.45\textwidth}
\centering
\includegraphics[scale=0.3]{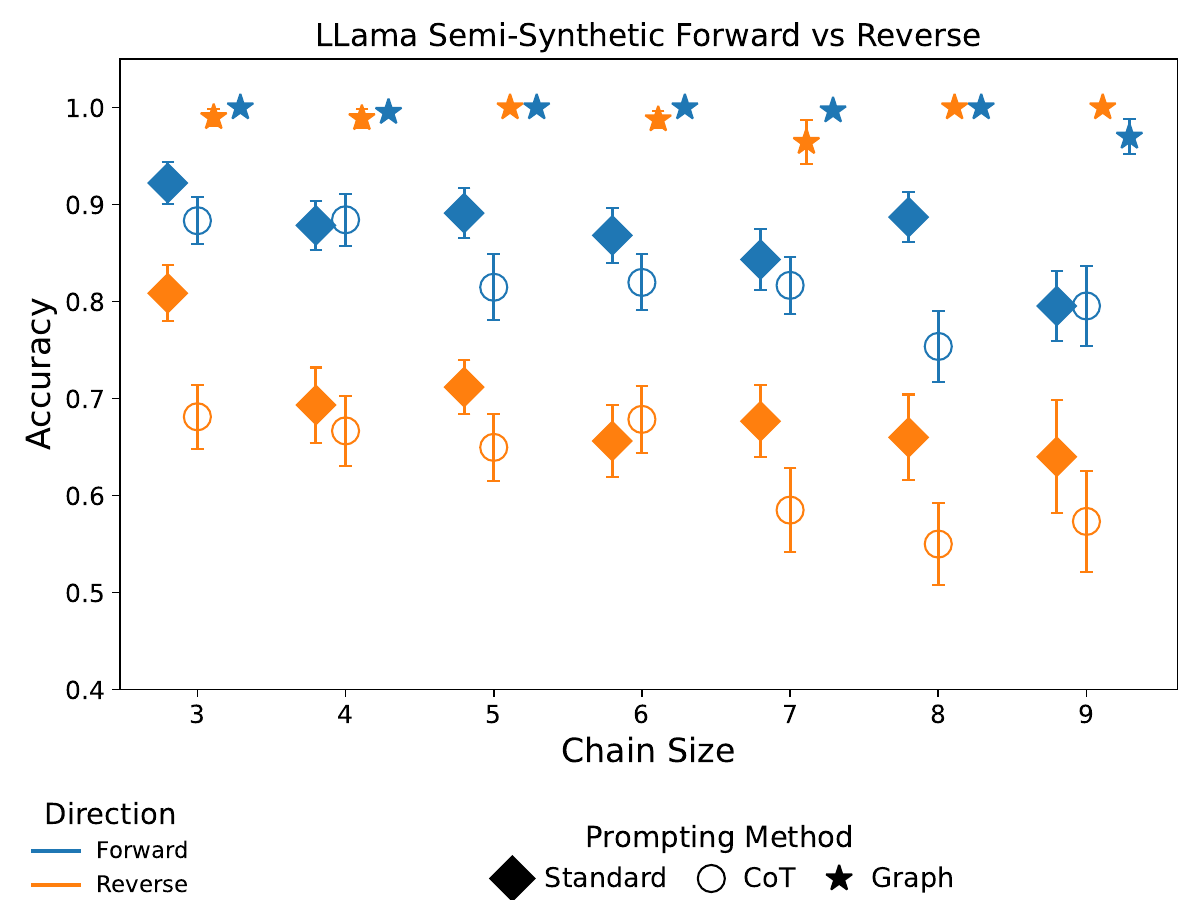}

\end{subfigure}
\hspace{10px}
\begin{subfigure}[b]{0.45\textwidth}
\centering
\includegraphics[scale=0.3]{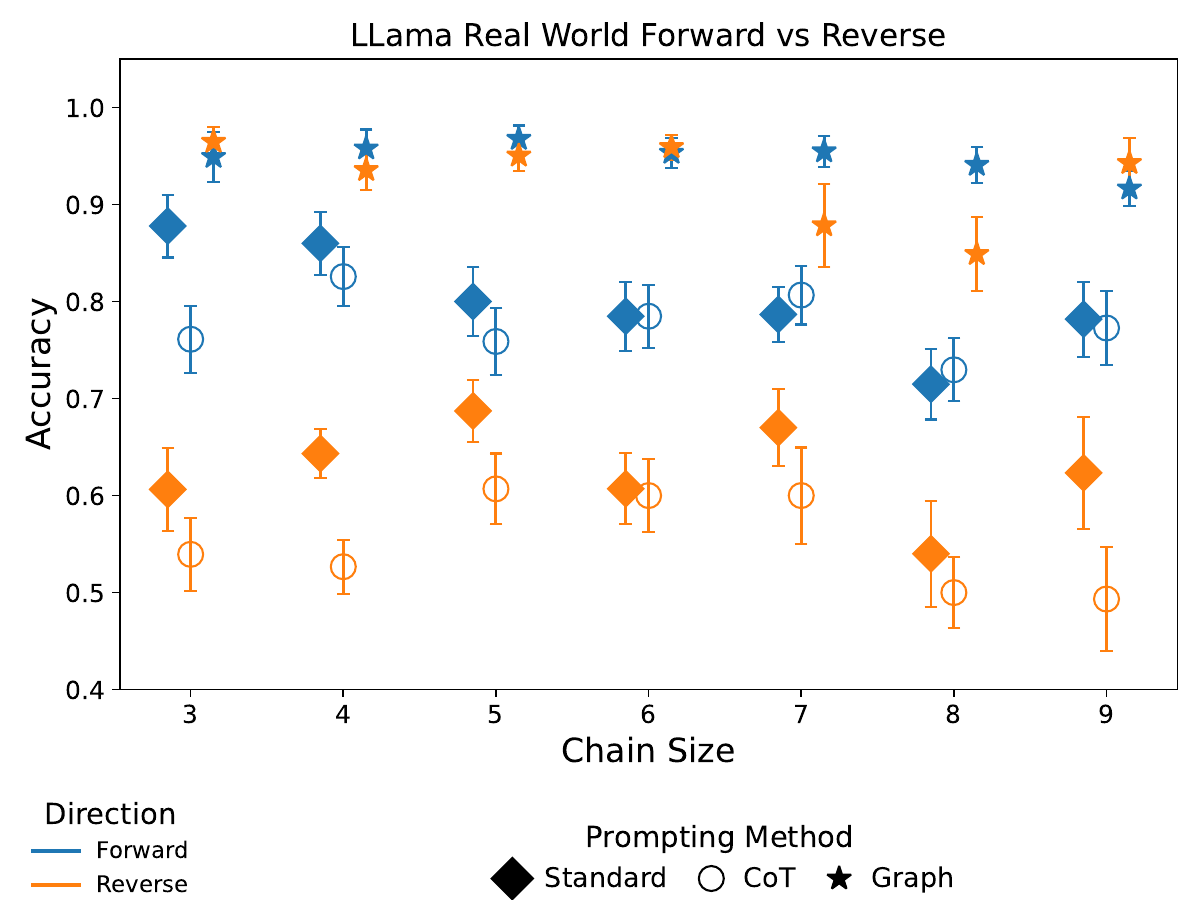}

\end{subfigure}
\caption{ (LLama 3.1 8B) The accuracy of various prompting strategies (error bars denote 95\% CIs) in the Semi-Synthetic and Real-World Regimes using CauseNet.
We observe that the accuracy is lower in the
reverse direction .
}
\end{figure*}

\subsection{Parametric Experiment LLama}

\begin{table*} [h!]
\centering
\begin{tabular}{@{}lccc@{}}
                 & \textbf{Standard} & \textbf{CoT} & \textbf{Graph}  \\
\toprule
\multicolumn{4}{l}{\textbf{Semi-synthetic}}                                                                      \\
\midrule
Without Conflict &  88.4                &    83.7           & 99.5                                  \\
With Conflict    & 61.4                              & 57.9                 & 98.2                  \\
\midrule
\multicolumn{4}{l}{\textbf{Real-world}}                                                                          \\
\midrule
Without Conflict & 81.6               & 79.2                & 95.1                                    \\
With Conflict    & 48.8               & 49.9                & 93.2                    \\    
\bottomrule
\end{tabular}
\caption{ (LLama 3.1 8B) The average accuracy across different narratives with the three prompting strategies partitioned by whether the cause-effect pairs
conflict with the LLM's parametric knowledge (we omit the 95\% CIs as they are smaller than $0.3$). 
}
\end{table*}

\subsection{Simple vs Complex LLama}
\begin{figure*}[h!]

    \begin{subfigure}[b]{0.45\textwidth}
\centering
\includegraphics[scale=0.3]{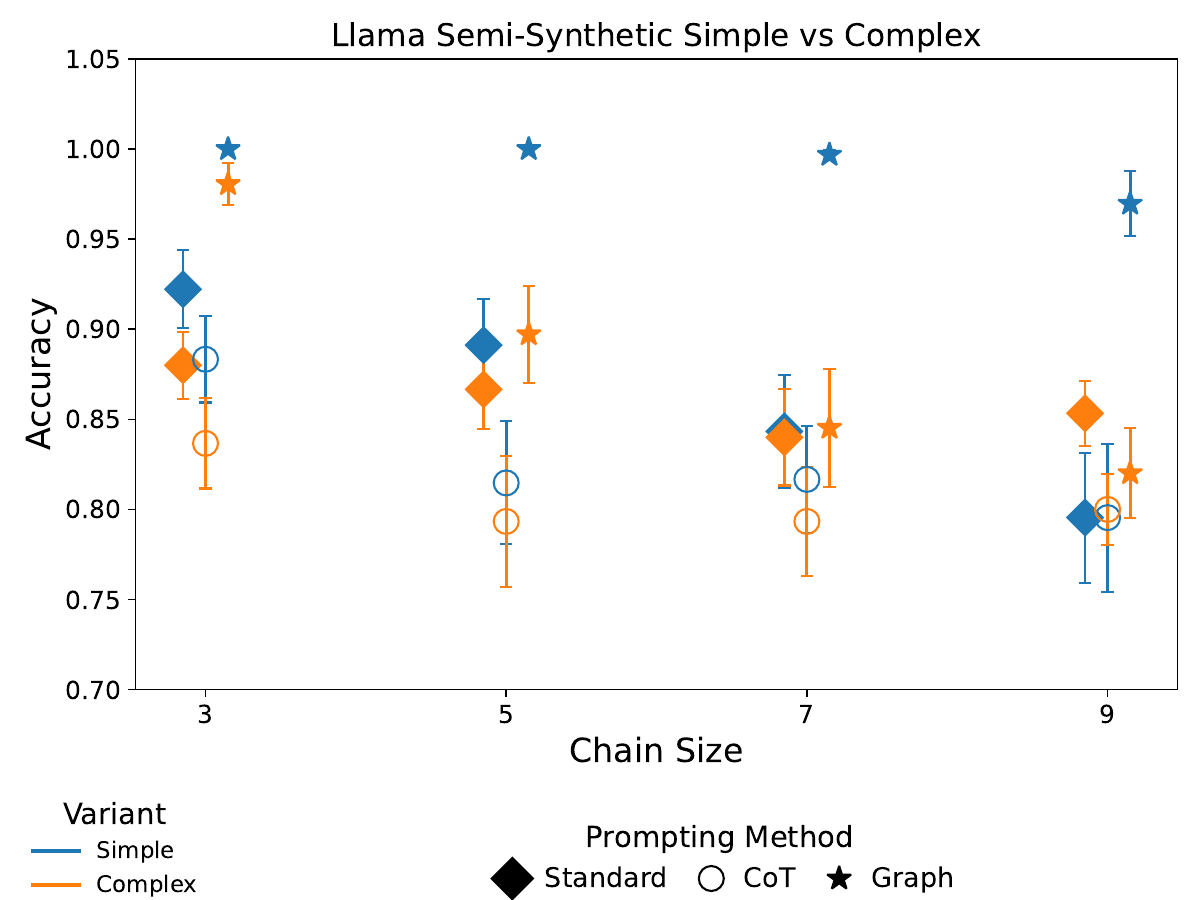}
\end{subfigure}
\hspace{10px}
\begin{subfigure}[b]{0.45\textwidth}
\centering
\includegraphics[scale=0.3]{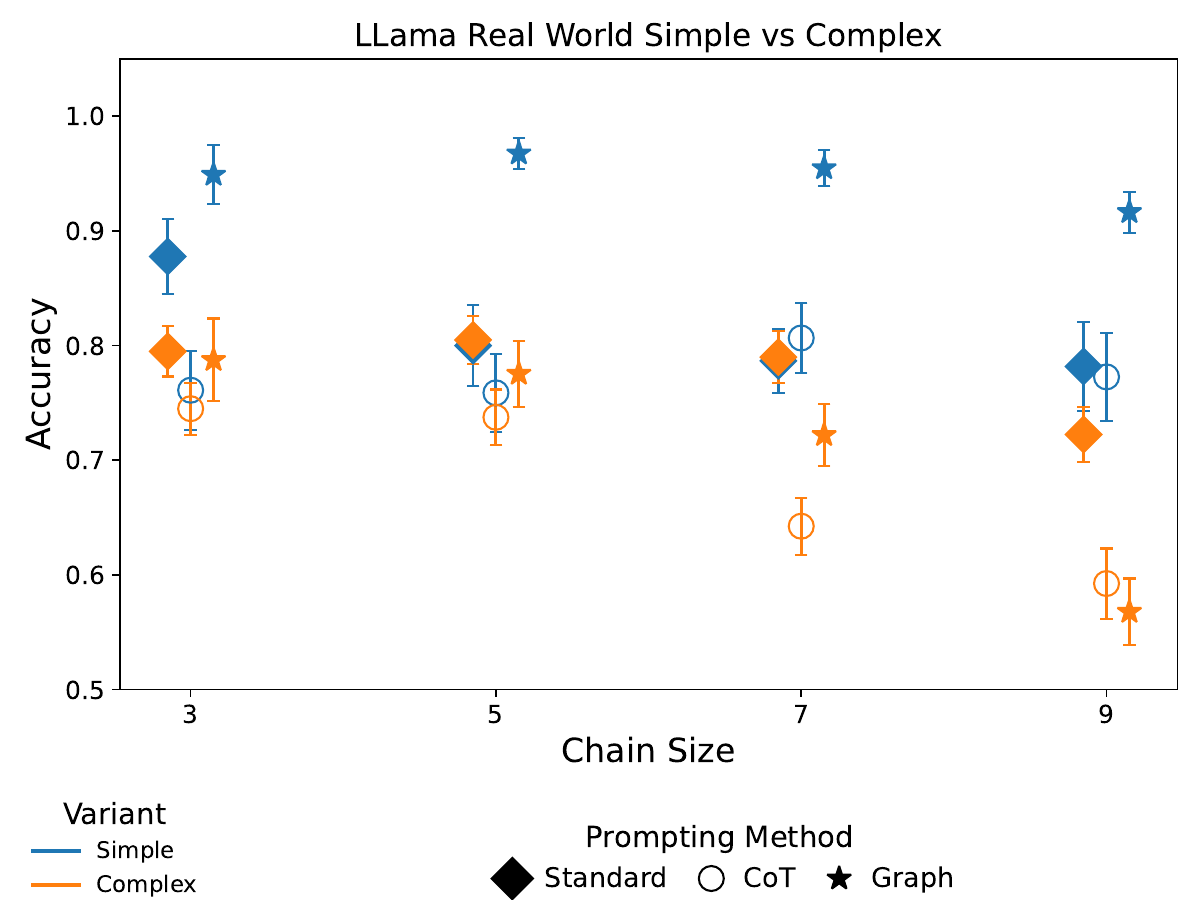}
\end{subfigure}

    \caption{ (LLama 3.1 8B) accuracy on narratives generated from Complex graphs as opposed to Simple chain graphs for semi-synthetic narratives (left) and real-world narratives (right).  95 $\%$ CI is shown. 
    }

\end{figure*}

\newpage

\end{document}